\documentclass[11pt]{article}

\usepackage[preprint]{acl}

\usepackage{times}
\usepackage{latexsym}

\usepackage[T1]{fontenc}

\usepackage[utf8]{inputenc}

\usepackage{microtype}

\usepackage{inconsolata}

\usepackage{graphicx}

\makeatletter
\newcommand*\myfontsize{
  \@setfontsize\myfontsize{7}{8}
}
\makeatother

\definecolor{myred}{rgb}{0.7, 0.3, 0.0}
\definecolor{myblue}{HTML}{054488}
\definecolor{mygreen}{HTML}{056b34}

\newcommand{\blue}[1]{\textbf{\textcolor{myblue}{#1}}}

\definecolor{darkgreen}{rgb}{0.0, 0.42, 0.24}
\usepackage{caption}
\usepackage{subcaption}
\usepackage{graphicx}
\usepackage{pifont}
\usepackage{titletoc}
\usepackage{amsfonts}
\usepackage{booktabs}
\usepackage{arydshln}
\usepackage{colortbl}
\usepackage{algorithm}
\usepackage{multirow}
\usepackage{wrapfig}
\usepackage[noend]{algpseudocode}
\usepackage{enumitem}
\usepackage{graphicx}
\usepackage{tcolorbox}
\usepackage{soul}
\usepackage{colortbl,array,xcolor}
\tcbuselibrary{breakable}
\definecolor{citecolor}{HTML}{0051f4}
\definecolor{pink}{HTML}{ed008c}
\hypersetup{
    colorlinks,
    linkcolor=citecolor,
    urlcolor=pink,
    citecolor=citecolor
}
\usepackage{hyperref}
\usepackage{url}
\usepackage{graphicx}
\usepackage{amsmath}

\title{
TourPlanner: A Competitive Consensus Framework with Constraint-Gated Reinforcement Learning for Travel Planning
}

\author{
Yinuo Wang$^{1 \dagger}$, Mining Tan$^{3,4 \dagger}$, Wenxiang Jiao$^1$, Xiaoxi Li$^{1,2}$, Hao Wang$^1$, \\
\textbf{Xuanyu Zhang$^1$, Yuan Lu$^1$, Weiming Dong$^{4,3*}$} \\
$^1$Xiaohongshu Inc. ~~ $^2$Renmin University of China\\
$^3$School of Artificial Intelligence, University of Chinese Academy of Sciences \\
$^4$MAIS, Institute of Automation, Chinese Academy of Sciences \\
\texttt{\{wangyinuo2, luyuan3\}@xiaohongshu.com}, \texttt{tanmining2024@ia.ac.cn} \\}
\begin{document}
\maketitle
\begin{abstract}

Travel planning is a sophisticated decision-making process that requires synthesizing multifaceted information to construct itineraries.
However, existing travel planning approaches face several challenges: (1) Pruning candidate points of interest (POIs) while maintaining a high recall rate; (2) A single reasoning path restricts the exploration capability within the feasible solution space for travel planning; (3) Simultaneously optimizing hard constraints and soft constraints remains a significant difficulty. To address these challenges, we propose TourPlanner, a comprehensive framework featuring multi-path reasoning and constraint-gated reinforcement learning.
Specifically, we first introduce a Personalized Recall and Spatial Optimization (PReSO) workflow to construct spatially-aware candidate POIs' set.
Subsequently, we propose Competitive consensus Chain-of-Thought (CCoT), a multi-path reasoning paradigm that improves the ability of exploring the feasible solution space.
To further refine the plan, we integrate a sigmoid-based gating mechanism into the reinforcement learning stage, which dynamically prioritizes soft-constraint satisfaction only after hard constraints are met.
Experimental results on travel planning benchmarks demonstrate that TourPlanner achieves state-of-the-art performance, significantly surpassing existing methods in both feasibility and user-preference alignment.

\end{abstract}

\section{Introduction}
\label{sec: introduction}
Travel planning is a sophisticated decision-making process that involves synthesizing large-scale multifaceted data, including accommodations, transportation, and points of interest (POIs), into an itinerary \citep{tang2024itinera, xie2024travelplanner, deng2025retail, shao2024chinatravel}. With the development of Large Language Models (LLMs) \citep{guo2025deepseek, achiam2023gpt, yang2025qwen3, comanici2025gemini} and reasoning technologies \citep{wei2022chain, yao2023tree, yu2025long}, agentic travel planning has gained prominence \citep{ning2025deeptravel, yang2025plan, wang2025triptailor, hao2025large}.
For instance, TravelPlanner \citep{xie2024travelplanner} and TripTailor \citep{wang2025triptailor} establish foundational benchmarks, employing methods such as ReAct \citep{zhang2024codeagent} and Reflection \citep{kambhampati2024llms} as baselines to construct itineraries. DeepTravel \citep{ning2025deeptravel} introduces a thinking-action-observation framework to iteratively generate plans.

Despite their success, there are still three main challenges for travel planning agents. First, the vast scale of candidate POIs often exceeds context-length limits, thereby compromising the quality of generated itineraries. Second, existing methods predominantly focus on constructing a single reasoning path; consequently, they often fail to adequately explore the solution space, leading to low feasibility of generated itineraries. Third, simultaneously satisfying hard constraints (e.g., valid visiting hours and non-repetitive POIs) and soft constraints (e.g., route efficiency and personalization) poses a significant challenge during the optimization process.

\begin{figure*}[t!]
  \centering
    \includegraphics[width=1\textwidth]{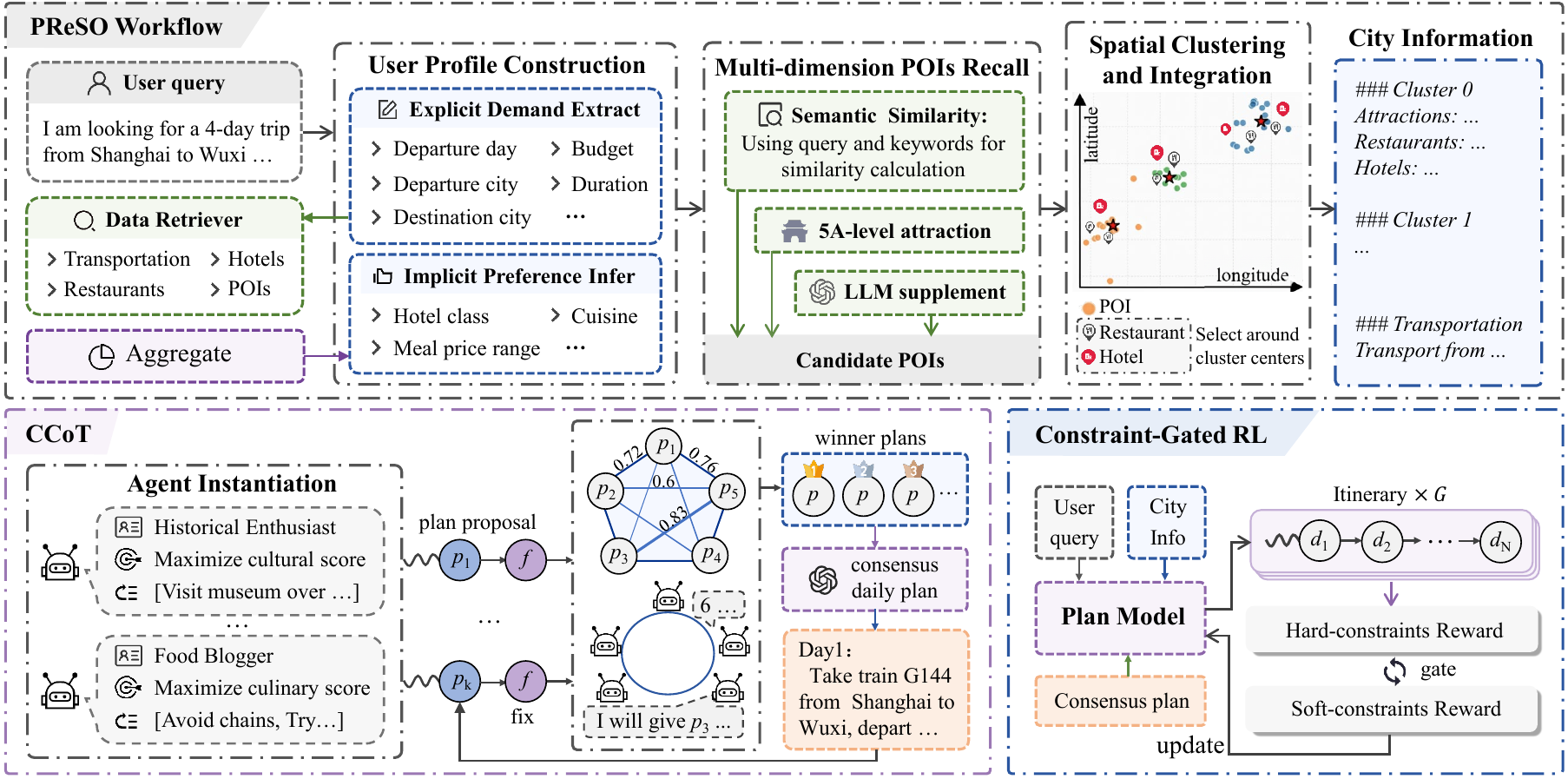}
  \caption{
     Overview of our TourPlanner Framework. Upon receiving a user query, the framework constructs  candidate information via the \textbf{PReSO} workflow to support the \textbf{CCoT} process. The CCoT involves three main phases: (1) Agent Instantiation, followed by iterative cycles of (2) Parallel Proposal Generation and (3) Competition and Consensus Arbitration for each of the $d$ days. During arbitration, the system synthesizes the top-$k$ proposals into a daily plan. Finally, a \textbf{Constraint-Gated RL} module refines the consensus itinerary to simultaneously optimize both soft constraints (personalization) and hard constraints (feasibility).
  }
  \label{fig: workflow}
\end{figure*}

In this work, we propose \textbf{TourPlanner}, a comprehensive framework designed to enhance the quality of generated itinerary. Specifically, we first design a workflow that incorporates a three-branch recall mechanism to effectively prune candidate POIs while maintaining a high recall rate, coupled with a clustering-based tagging approach to integrate contextual information. Furthermore, we introduce a multi-path reasoning method to improve exploration capabilities within the solution space, resulting in highly feasible itineraries. Finally, we propose a sigmoid-based gating mechanism to address the challenge of simultaneously optimizing both hard and soft constraints. Our key contributions are summarized as follows:
\begin{itemize}[leftmargin=10pt]
    \item \textbf{Personalized Recall and Spatial Optimization (PReSO):}
    We design a preprocessing workflow to prune candidate POIs. This workflow extracts explicit demands and infers implicit user preferences to guide a multi-dimensional recall mechanism. By employing spatial clustering, it anchors accommodations and dining options around POI centroids, formulating a spatially compact set of candidate POIs. These filtered candidate POIs, enriched with cluster category information, serve as input for the LLM.
    \item \textbf{Competitive consensus Chain-of-Thought (CCoT):}
    We introduce a multi-path reasoning paradigm that enhances the capability of exploring the solution space. By instantiating specialized agents with distinct personas, the system generates parallel daily proposals. These proposals undergo a three-phase arbitration protocol—including Proposal Diversity Weighting, Parallel Peer Review, and Weighted Consensus Selection—to resolve multi-objective conflicts, ensuring the final itinerary reflects a balanced, expert-level consensus, thereby significantly enhancing itinerary feasibility.
    \item \textbf{Constraint-Gated Reinforcement Learning for Plan Refinement:}
    Recognizing that consensus plans may still fall short in simultaneously optimizing hard constraints and soft constraints, we introduce a sigmoid-based gating mechanism within the reinforcement learning (RL) refinement stage. By adopting a curriculum-like approach, this mechanism dynamically increases the weight of soft-constraint rewards only upon the satisfaction of hard constraints.
\end{itemize}

\section{Related Work}
\label{sec: related work}

\paragraph{LLMs for Travel Planning.}
Recent advances in LLMs have profoundly reshaped travel planning, enabling natural-language interaction and multi-constraint reasoning for itinerary generation. Early research primarily focused on constructing benchmarks to evaluate the planning capabilities. TravelPlanner~\citep{xie2024travelplanner} introduced the first large-scale sandbox benchmark with extensive travel data and tool access, revealing that existing LLMs still struggle to achieve high task success rates due to weak grounding and limited constraint-handling, despite their strong reasoning abilities. TripTailor~\citep{wang2025triptailor} further advanced real-world benchmarking by incorporating over 500,000 POIs and authentic user itineraries, thereby facilitating large-scale evaluation of personalization and rationality.

Parallel to these developments, hybrid frameworks have emerged to address the unreliability of purely LLM-based planning. Hao et al.~\citep{hao2025large} integrated satisfiability solvers for constraint validation, substantially improving success rates, while TRIP-PAL~\citep{de2024trip} combined automated planners with LLM reasoning to ensure constraint satisfaction. These methods demonstrate that formal modules can significantly enhance the reliability of generated plans.

\begin{figure}[t!]
 \centering
   \includegraphics[width=\columnwidth]{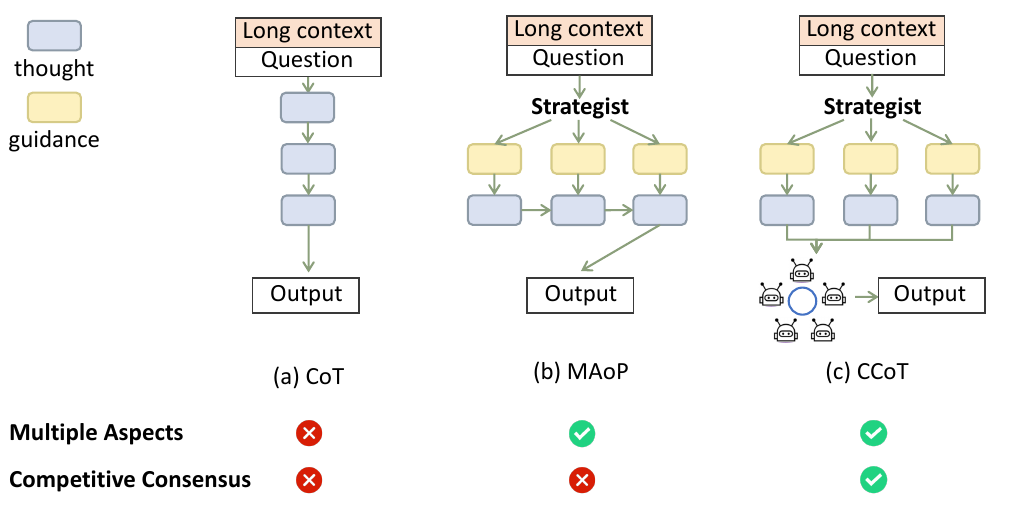}
 \caption{Comparison between prior methods.
 }
 \label{fig: compare}
\end{figure}

Subsequent research has proposed LLM-based travel planning agents capable of dynamic tool interaction and contextual reasoning. TravelAgent~\citep{chen2024travelagent} introduced a modular system comprising tool-use, recommendation, planning, and memory components to generate personalized itineraries in dynamic environments. Yang et al.~\citep{yang2025plan} developed wide-horizon reasoning through Multiple Aspects of Planning (MAoP) and simulated evaluation environments, while RETAIL~\citep{deng2025retail} constructed a topic-guided multi-agent framework designed to manage implicit user intent and environmental constraints, thereby revealing persistent challenges in generalization. In contrast to the above methods that focus on a single reasoning path, we propose TourPlanner, a method that generates multiple parallel, competing sub-optimal strategies and reaches a consensus through a simulated round-table discussion as shown in Figure~\ref{fig: compare}.

\paragraph{Reinforcement Learning for LLMs.}
RL has emerged as a key paradigm for enhancing the capability of models~\citep{schulman2017proximal, haarnoja2018soft, ren2024diffusion, wang2024diffusion}. While early approaches like RLHF~\citep{ouyang2022training} aligned models via human preferences, recent methods such as DPO~\citep{rafailov2023direct}, GRPO~\citep{shao2024deepseekmath}, DAPO~\citep{yu2025dapo}, and GSPO~\citep{zheng2025group} have further optimized training stability and reasoning at scale. These advancements underscore RL's effectiveness in complex decision-making tasks, making it a powerful tool for travel planning.

\section{TourPlanner}
\label{sec: method}
\subsection{Task Description}
This work focuses on the single-turn travel itinerary generation problem, i.e., generating a complete travel itinerary ($I$) from a single user query ($Q$). The user query is expressed in natural language form containing explicit user requirements (e.g., departure city, arrival city, departure date, arrival date, etc.). The travel plan consists of three components: accommodation arrangements, transportation logistics, and a detailed daily schedule.
The primary symbol definitions and LLM prompts are summarized in Appendices~\ref{app: symbol} and~\ref{app: prompts}, respectively.

\subsection{Personalized Recall and Spatial Optimization}
\label{workflow}
The vast number of candidate POIs presents significant challenges for LLMs in travel planning, potentially exceeding context-length limits and diminishing output quality. To address this issue, we propose the \textbf{Personalized Recall and Spatial Optimization (PReSO)} workflow for effective pre-filtering before travel planning. This includes three steps: (1) \textit{User Profile Construction}, to construct a comprehensive user profile by extracting explicit requirements and inferring implicit preferences; (2) \textit{Multi-dimension POIs Recall}, to ensure the coverage of the most relevant POIs; (3) \textit{Spatial Clustering and Integration}, to address the geographical dispersion of initially retrieved POIs. The overall workflow is illustrated in Figure~\ref{fig: workflow}.

\paragraph{User Profile Construction.}
Existing approaches for user profile construction primarily rely on explicit user requirements (e.g., departure/return dates, trip duration, destination, and budget). However, user queries often contain valuable implicit preferences, such as preferred hotel class, meal price range, or restaurant type, that are not directly stated but can be inferred. To capture these latent signals, we augment the raw user query with city-specific statistical data and leverage an LLM to extrapolate implicit needs. The resulting comprehensive profile, which synthesizes both explicit requirements and inferred preferences, subsequently informs the filtering and planning stages.

\paragraph{Multi-dimension POIs Recall.}
POI Recall quantifies the proportion of ground-truth POIs successfully retrieved by the filtering process. The core challenge is to effectively select the most relevant POIs for a trip from a massive candidate pool. Inspired by the principles of \textit{multi-channel retrieval} in recommendation systems~\citep{huang2024comprehensive}, we propose a three-way recall mechanism. First, we employ an embedding model~\citep{xiao2024c} to identify relevant POIs based on the semantic similarity. Beyond sentence-level user preferences, we extract keywords and expand them with synonyms to enhance the analysis. Second, since travelers usually expect to visit the most renowned landmarks of a city, we recall attractions rated 4A or above to ensure the inclusion of canonical highlights. Candidate attractions are ranked based on popularity and user ratings, with priority given to those that are both widely visited and highly rated. Third, an LLM is leveraged to supplement the candidate pool by identifying and suggesting attractions that align with user preferences. This hybrid retrieval mechanism yields a final candidate set that is both semantically rich and representatively complete, providing a high-quality initialization for subsequent route optimization.

\paragraph{Spatial Clustering and POI Integration.} Route efficiency, measured by the total travel distance required to connect all specified POIs, is a critical metric for evaluating the rationality of real-world travel itineraries. However, naively retrieved POIs are often geographically dispersed, leading to inefficient and impractical travel itineraries. To address this challenge, we cluster attractions based on their geographical coordinates using DBSCAN \citep{ester1996density}, a density-based method with adaptive $\varepsilon$-neighborhood adjustment. The resulting cluster centroids serve as crucial spatial anchors for guiding the selection of nearby accommodations and restaurants. Subsequently, the generated cluster labels are integrated as an additional attribute into the information for accommodations, restaurants, and POIs, constituting clustered urban data for use in the planning stage. This spatial integration ensures that the final itineraries are geographically compact while optimally satisfying user requirements and preferences.

\subsection{Competitive consensus Chain-of-Thought}
\label{sec: ccot}
Existing agentic travel planning methods, which typically employ long-horizon or wide-horizon CoT~\citep{wei2022chain, yao2023tree, yang2025plan}, remain constrained to a single reasoning path. This limitation constraints their exploration capability within the solution space, making them struggle to resolve the multi-objective conflicts (e.g., maximizing cultural value while simultaneously minimizing cost). To mitigate this challenge, we propose the Competitive consensus Chain-of-Thought (CCoT). CCoT shifts the planning paradigm from a single-path reasoning to a multi-path reasoning. It explicitly models diverse user needs as specialized reasoning agents and resolves their conflicts via a weighted-consensus mechanism, thereby enabling a balance among conflicting objectives. The CCoT framework operates iteratively for each day of the trip, executing three core stages: (1) \textit{Agent Instantiation}, (2) \textit{Parallel Proposal Generation}, and (3) \textit{Competition and Consensus Arbitration}.

\subsubsection{Agent Instantiation}
Given a user query $Q$, a static set of $N$ specialized reasoning agents $\mathcal{A} = \{A_1, A_2, \ldots, A_N\}$ is initialized. This agent team is maintained consistently across the entire planning horizon. Each agent $A_i$ is rigorously defined by a distinct Identity ($I_i$), a measurable Objective ($O_i$), and a set of Ranked Priorities ($R_i$).
For instance, consider a user query prioritizing ``Culture, Gourmet, and limited Budget''. This instantiation yields the following agents:
\begin{itemize}[leftmargin=10pt]
    \item $A_{\text{Culture}}$. $I_C$: Historical Enthusiast; $O_C$: Maximize Cultural Experience Score; $R_C$: [World Heritage Sites, Allocate $\ge 3$ hours for museums, Historic Districts].
    \item $A_{\text{Gourmet}}$. $I_G$: Food Blogger; $O_G$: Maximize Local Culinary Satisfaction Score; $R_G$: [Sample Regional Specialties, Avoid Chain Establishments].
    \item $A_{\text{Budget}}$. $I_B$: Fiscally Conservative Manager; $O_B$:
    Minimize Total Expenditure; $R_B$: [Utilize Economy Transportation, Select Economy Dining Options].
\end{itemize}
This modular instantiation ensures that every diverse user preference is explicitly represented and prepared to participate in the subsequent competition and consensus arbitration processes.

\subsubsection{Parallel Proposal Generation}
The process of proposal generation follows a \textbf{Skeleton-then-Refine} paradigm. First, a \textit{General Expert Agent} generates a base routing skeleton $P_{\text{base}, d}$ for day $d$. Subsequently, each instantiated agent $A_i$ independently modifies this skeleton to produce a refined daily itinerary proposal $P_{i, d}$, which is strictly optimized to maximize its own objective $O_i$. Then, any attraction, restaurant, or accommodation already included in the consensus plan $C_{1..d-1}$ is removed from the planning context, resulting in the updated given information $G_d$. Formally, the proposal generation process for agent $i$ on day $d$ is defined as:
\begin{equation}
    P_{i, d} \leftarrow A_i(I_i, O_i, R_i, Q, G_d, P_{\text{base}, d}, C_{1..d-1}).
\end{equation}
This parallel exploration strategy ensures a wide-horizon search across diverse solution spaces, generating a set of robust, competing daily plans. Following generation, all proposals are subjected to rule validation to guarantee that the subsequent stages operate on reliable input.

\subsubsection{Competition and Consensus Arbitration}
This is the core stage of CCoT, which is dedicated to resolving the conflicts among the $N$ parallel itinerary proposals.

Instead of relying on a direct fusion, we implement a \textbf{three-phase arbitration protocol} to generate the final consensus plan $C_d$.

\paragraph{Proposal Diversity Weighting.}
To ensure the final plan incorporates diverse specialized insights, we compute a diversity weight $w_i$ for each agent. The proposal $P_i$ is embedded into a vector $\mathbf{e}_i$, forming an $N \times N$ similarity matrix $\mathbf{S}$ with entries $\mathbf{S}_{ij} = \cos(\mathbf{e}_i, \mathbf{e}_j)$.
For each agent $i$, we calculate its average similarity to peers, $\bar{S}_i = \frac{1}{N-1} \sum_{j \neq i} \mathbf{S}_{ij}$. The raw weight $\hat{w}_i$ is set inversely proportional to $\bar{S}_i$ to reward uniqueness, and then normalized:
\begin{equation}
    \hat{w}_i = \frac{1}{\bar{S}_i + \epsilon}, \quad w_i = \frac{\hat{w}_i}{\sum_{k=1}^N \hat{w}_k},
\end{equation}
where $\epsilon$ is a smoothing constant. This formulation ensures $w_i$ effectively measures proposal diversity, preventing the consensus from collapsing into a set of near-identical candidates.

\paragraph{Parallel Proposal Review.}
In this phase, all proposals undergo peer review before final selection. Each agent $A_i$ acts as a reviewer, assigning a numerical quality score $s_{i,j}$ and a natural language critique $T_{i,j}$ to every competing proposal $P_j$. The scoring is based on the agent's specific objective $O_i$, ranked priorities $R_i$ and proposal feasibility. This parallel scoring process generates an $N \times N$ score matrix that explicitly captures the multi-objective trade-offs and feasibility as perceived by the specialized agents.

\paragraph{Weighted Consensus Selection.}
Finally, the system functions as an arbiter, leveraging diversity weights $w_i$ and peer review scores $s_{i,j}$ to derive a consensus. An aggregated score for each proposal $P_j$ is computed as the weighted summation of peer evaluations:
\begin{equation}
    \text{Score}(P_j) = \sum_{i=1}^N w_i \cdot s_{i,j}.
\end{equation}
The top-$k$ proposals with the highest scores are selected as candidates for day $d$. These candidates are subsequently synthesized into a unified daily plan by an LLM, utilizing specific constraints as the system prompt and the $k$ proposals along with their qualitative peer insights $\{T_{i,j}\}$ as the user prompt. This configuration allows the model to effectively integrate the strengths of diverse proposals while mitigating their respective weaknesses. Finally, the generated itinerary is appended to the cumulative schedule $C_{1..d-1}$, updating the overall plan to $C_{1..d}$. This structured arbitration protocol establishes a robust mechanism for resolving multi-objective conflicts, ensuring the consistent generation of high-quality travel plans.

\subsection{Constraint-Gated Reinforcement Learning}
Despite the multi-path deliberation facilitated by CCoT, the resulting itineraries may still fail to achieve optimal alignment with user preferences. Consequently, we propose a supplementary refinement phase, employing a constraint-gated reinforcement learning (RL) method to further optimize the plan.

\paragraph{Optimizing Challenge.}
The reward function for travel planning tasks can be categorized into hard-constraint and soft-constraint rewards, both of which are essential for the overall quality of the generated itineraries. Specifically, rewards for hard constraints—such as the absence of hallucinations, valid visiting hours, and non-repetitive POIs—are typically sparse and binary, yet fundamental to plan feasibility. In contrast, rewards for soft constraints, including budget reasonableness, route efficiency, and personalization, are generally dense and continuous. As demonstrated in Table~\ref{tab: ablation_agent_number}, a naive additive reward function ($R = R_{\text{hard}} + R_{\text{soft}}$), commonly used in Vanilla RL, fails to simultaneously optimize both hard and soft constraints, leading to a significant reduction in the final pass rate of the generated plans.

\paragraph{Constraint-Gated Reward.}
To address this challenge, we introduce a sigmoid-based gating mechanism. The total reward function is designed as follows, featuring a factor $\alpha(\eta)$:
\begin{equation}
R = R_\text{hard} + \alpha(\eta) \cdot R_\text{soft},
\end{equation}
where $\alpha(\eta)$ is defined as $\alpha(\eta)=\frac{1}{1 + e^{-k(\eta - \tau)}} \in [0, 1]$.
This mechanism dynamically modulates the optimization objective based on hard constraint satisfaction ($\eta$):
(1) \textit{Hard Constraint Focus}: When the hard constraint score $\eta$ is below the threshold $\tau$, the scaling factor $\alpha \rightarrow 0$. This effectively masks the $R_{\text{soft}}$ signal ($R \approx R_{\text{hard}}$), compelling the agent to focus exclusively on resolving hard constraint violations.
(2) \textit{Quality Enhancement}: Once hard constraints are met ($\eta \geq \tau$), $\alpha$ rapidly increases towards 1, transitioning the objective to $R \approx R_{\text{hard}} + R_{\text{soft}}$. This smooth transition inspired by the curriculum learning, enabling both hard and soft constraints to be optimized and satisfied simultaneously.

\paragraph{Optimization Strategy.}
Following the Group Sequence Policy Optimization~(GSPO) framework \citep{zheng2025group}, we sample data from the training dataset $\mathcal{D}$, roll out $G$ trajectories for each query, and optimize the agent according to the objective function:
\begin{equation}
\resizebox{1.0\hsize}{!}{
$
\begin{aligned}
\mathcal{J}_{\mathrm{GSPO}}(\theta)
&= \mathbb{E}_{x, \{y_i\} \sim \pi_{\text{old}}} \Big[ \\
&\quad \frac{1}{G} \sum_{i=1}^{G}
\min\left(
r_i(\theta)\hat{A}_i,\,
\text{clip}(r_i(\theta), 1\text{-}\epsilon, 1\text{+}\epsilon)\hat{A}_i
\right)
\Big],
\end{aligned}
$
}
\label{eq: gspo}
\end{equation}
where $\hat{A}_i = \frac{r(x, y_i) - \mathrm{mean}(\{r(x, y_j)\}_{j=1}^{G})} {\mathrm{std}(\{r(x, y_j)\}_{j=1}^{G})}$ represents the group-based advantage estimation, $\varepsilon$ is the clipping parameter, and $s_i(\theta)$ denotes the importance ratio. Further technical details of the algorithm are provided in Appendix \ref{appendix: gspo}.

\section{Experiments}
\label{sec: experiments}

\subsection{Experimental Settings}

\paragraph{Task and Dataset.}
A realistic and stable experimental environment is essential for systematically evaluating the performance of travel planning systems. To ensure reproducible and consistent benchmarking, we adopt the \textbf{TripTailor} sandbox~\citep{wang2025triptailor}, a large-scale simulation environment encompassing 40 major Chinese cities. Detailed descriptions of TripTailor are provided in Appendix~\ref{app: triptailor}.

\paragraph{Evaluation Metrics.}
We employ five primary metrics to evaluate the performance of travel planning models. Detailed formulations for all metrics are provided in Appendix~\ref{app: metrics}.

\paragraph{Compared Methods.}
Our evaluation incorporates a comprehensive set of LLM backbones and planning methodologies. Detailed descriptions of these baselines are provided in Appendix~\ref{app: baselines}.

\paragraph{Reward Setting.}
The reward function comprises two components: a hard-constraint reward ($R_{\text{hard}}$) and a soft-constraint reward ($R_{\text{soft}}$). The hard-constraint reward enforces fundamental feasibility by penalizing violations such as hallucinations, incomplete information, invalid visiting hours, and repetitive POIs. The soft-constraint reward incentivizes plan quality, focusing on metrics like budget reasonableness, route efficiency, and alignment with the user preference reward model. Detailed mathematical formulations for these reward components are provided in Appendix~\ref{app: reward}.

\begin{table*}[!t]
\centering
\caption{
Results on the TripTailor benchmark across various planning approaches and LLM backbones. The best results are highlighted in bold, while the second-best results are \underline{underlined}.
}
\setlength\tabcolsep{5pt}
\fontsize{8.1pt}{10.5pt}\selectfont
\begin{tabular}{lccccccc}
\toprule
\multirow{2}{*}{\textbf{Method}}
& \multicolumn{2}{c}{\shortstack{\textbf{Feasibility}\\\textbf{Pass Rate ($\uparrow$)}}}
& \multicolumn{2}{c}{\shortstack{\textbf{Rationality}\\\textbf{Pass Rate ($\uparrow$)}}}
& \multirow{2}{*}{\shortstack{\textbf{Average Route}\\\textbf{Distance Ratio}  ($\downarrow$)}}
& \multirow{2}{*}{\shortstack{\textbf{Final}\\\textbf{Pass Rate}  ($\uparrow$)}}
& \multirow{2}{*}{\shortstack{\textbf{Final}\\\textbf{Surpassing Rate}  ($\uparrow$)}} \\
\cmidrule(lr){2-3}
\cmidrule(lr){4-5}
& Micro & Macro & Micro & Macro &  &  \\
\midrule
\multicolumn{7}{l}{\textit{\textbf{Direct Planning}}} \\
GPT-4o & 95.4 & 90.7 & 73.3 & 21.9 & 5.98 & 6.7 & 3.7 \\
Qwen3-235B-A22B-Instruct & 81.5 & 64.6 & 78.1 & 29.7 & 2.92 & 11.0 & 9.0 \\
Qwen3-30B-A3B-Thinking & 84.6 & 69.7 & 74.0 & 23.2 & 2.51 & 10.8 & 6.1\\
DeepSeek-R1 & 91.8 & 86.3 & 79.5 & 30.7 & 2.58 & 14.8 &  10.6\\
\midrule
\multicolumn{7}{l}{\textit{\textbf{ReAct Planning}}} \\
GPT-4o & 93.1 & 90.7 & 74.4 & 21.4 & 3.65 & 9.4 & 6.4\\
Qwen3-235B-A22B-Instruct & 85.4 & 78.2 & 73.0 & 17.5 & 3.42 & 6.2 & 4.6 \\
Qwen3-30B-A3B-Thinking & 87.2 & 81.2 & 65.8 & 15.2 & 5.00 & 5.6 & 3.3\\
DeepSeek-R1 & 89.8 & 87.5 & 76.4 & 23.8 & 3.19 & 12.8 & 9.0\\
\midrule
\multicolumn{7}{l}{\textit{\textbf{TripTailor Workflow}}} \\
GPT4o & \underline{99.4} & \underline{98.8} & 83.8 & 31.3 & 5.64 & 14.1 & 8.7\\
Qwen3-235B-A22B-Instruct & 95.4 & 92.2 & 82.0 & 25.7 & 4.27 & 10.2 & 6.0 \\
Qwen3-30B-A3B-Thinking & 97.8 & 95.7 & 88.0 & 47.3 & 4.84 & 23.1 & 3.7\\
DeepSeek-R1 & 97.9 & 96.3 & 89.3 & 48.6 & 4.11 & 26.9 & 14.3\\
\midrule
\multicolumn{7}{l}{\textit{\textbf{TourPlanner w/o RL (Ours)}}} \\
GPT4o & \textbf{100.0} & \textbf{100.0} & \textbf{98.2} & \underline{91.5} & 2.28 & 53.7 & \underline{26.3}\\
Qwen3-235B-A22B-Instruct & \textbf{100.0} & \textbf{100.0}  & 97.8 & 89.4 & \underline{2.21} & 55.4 & 24.1\\
Qwen3-30B-A3B-Thinking & \textbf{100.0} & \textbf{100.0}  & 98.0 & 90.5 & 2.26 & 54.3 & 20.1\\
DeepSeek-R1 & \textbf{100.0} & \textbf{100.0} & \underline{98.1} & \textbf{91.8} & 2.23 & \underline{55.8} & 25.1\\
\midrule
\multicolumn{7}{l}{\textit{\textbf{TourPlanner (Ours)}}} \\
\rowcolor[RGB]{236,244,252}
Constraint-Gated RL & \textbf{100.0} & \textbf{100.0} & 97.1 & 88.7 & \textbf{2.15} & \textbf{56.1} & \textbf{30.2} \\
\bottomrule
\label{tab: main_results}
\end{tabular}
\end{table*}

\begin{figure}
    \centering
    \includegraphics[width=\linewidth]{ 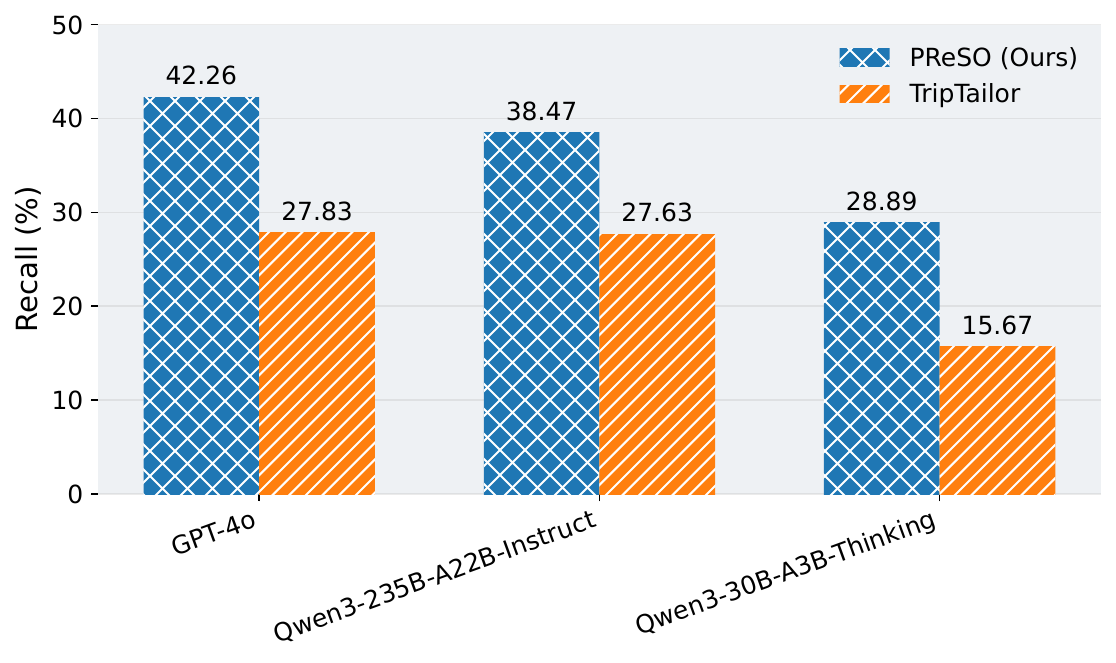}
    \caption{
    Recall Performance of PReSO versus TripTailor. Our workflow consistently outperforms TripTailor across various LLMs, achieving a higher proportion of identified ground-truth travel elements.
    }
    \label{fig:ablation_recall}
\end{figure}

\begin{table*}[!t]
\centering
\caption{Ablation study on the TripTailor benchmark. We analyze the impact of varying the number of agents using Qwen3-235B-A22B-Instruct and evaluate different refinement methods applied to the initial consensus plan generated by TourPlanner. The RL refinement utilizes Qwen3-30B-A3B-Instruct as the base model. Bold values indicate the best performance within each ablation group.}
\setlength\tabcolsep{4pt}
\fontsize{8.1pt}{10.5pt}\selectfont
\begin{tabular}{lccccccc}
\toprule
\multirow{2}{*}{\textbf{Method}}
& \multicolumn{2}{c}{\shortstack{\textbf{Feasibility}\\\textbf{Pass Rate ($\uparrow$)}}}
& \multicolumn{2}{c}{\shortstack{\textbf{Rationality}\\\textbf{Pass Rate ($\uparrow$)}}}
& \multirow{2}{*}{\shortstack{\textbf{Average Route}\\\textbf{Distance Ratio}  ($\downarrow$)}}
& \multirow{2}{*}{\shortstack{\textbf{Final}\\\textbf{Pass Rate}  ($\uparrow$)}}
& \multirow{2}{*}{\shortstack{\textbf{Final}\\\textbf{Surpassing Rate}  ($\uparrow$)}} \\
\cmidrule(lr){2-3}
\cmidrule(lr){4-5}
& Micro & Macro & Micro & Macro &  &  \\
\midrule
\multicolumn{7}{l}{\textit{\textbf{TourPlanner}}} \\
Default~(4-6 agents) & \textbf{100.0} & \textbf{100.0} & \textbf{97.8} & 89.4 & 2.21 & \textbf{55.4} & 24.1 \\
10 agents & \textbf{100.0} & \textbf{100.0} & 97.2 & \textbf{90.1} & 2.25 & 54.8 & \textbf{24.4} \\
3 agents & \textbf{100.0} & \textbf{100.0} & 97.0 & 87.3 & 2.31 & 51.9 &  21.6\\
Direct Combine w/o CCoT & \textbf{100.0} & \textbf{100.0} & 95.3 & 84.9 & \textbf{2.18} & 47.8 & 18.7\\
\midrule
\multicolumn{7}{l}{\textit{\textbf{TourPlanner}}} \\
Constraint-Gated RL & \textbf{100.0} & \textbf{100.0} & \textbf{97.1} & \textbf{88.7} & 2.15 & \textbf{56.1} & \textbf{30.2} \\
Vanilla RL & 99.3 & 99.1 & 91.8 & 67.9 & \textbf{1.91} & 47.5 & 29.3 \\
Direct Refine w/o RL & 96.5 & 93.4 & 95.5 & 81.3 & 2.59 & 47.1 & 29.0 \\
\bottomrule
\label{tab: ablation_agent_number}
\end{tabular}
\end{table*}

\subsection{Main Results}
\paragraph{Achieve Superior Performance in Multi-Constraint Satisfaction.}
Table~\ref{tab: main_results} lists the results on the TripTailor benchmark. One of the most significant findings is the dramatic performance leap in Macro Rationality, which measures an agent's ability to satisfy all complex travel constraints simultaneously. While baseline methods (\textit{Direct} and \textit{ReAct}) achieve reasonable results on individual Micro metrics, they struggle to balance multiple requirements, often resulting in Macro Pass Rates below 30\%. In contrast, our method exceeds 88\% across all LLM backbones, representing a substantial gain over baseline methods. This suggests that our method effectively overcomes the ``all-or-nothing'' challenge inherent in complex, high-dimensional planning tasks. A concrete example is provided in Appendix~\ref{app: case}.

\paragraph{Effectively Optimize Spatial Efficiency and Route Coherence.}
The results demonstrate that our method substantially improves the Average Route Distance Ratio, reducing it from as high as 5.98 (e.g., \textit{Direct Planning} with GPT-4o) to 2.15. Unlike traditional planning approaches that often generate spatially illogical sequences, our framework consistently produces itineraries with transport efficiency. This indicates that the agent is not merely selecting valid POIs but is actively optimizing the spatial-temporal flow of the journey.

\paragraph{Ensure Robust Feasibility and Model-Agnostic Generalization.}
Across all tested LLM backbones, TourPlanner achieves a perfect 100\% Feasibility Pass Rate, effectively eliminating the hallucinations and sandbox mismatches that plague traditional agents. This success, combined with a Final Surpassing Rate consistently exceeding 20\% and outperforming all baseline methods, suggests that the performance gains are driven by the TourPlanner framework itself rather than the specific parameters of a single model. Furthermore, after refinement by the RL-trained model, our method, achieves a Final Pass Rate of 56.1\% and a Final Surpassing Rate of 30.2\%, outperforming the initial consensus plan.

\subsection{Analysis}
\paragraph{Effectiveness of the PReSO Workflow.}
To verify the effectiveness of PReSO workflow, we conducted a comparative analysis against the TripTailor baseline, specifically evaluating the recall rate of candidate items. In this evaluation, recall measures the ability of workflow to successfully identify ground-truth hotels, restaurants, and POIs within the sandbox environment. We benchmarked both workflows across three backbone models: GPT-4o, Qwen3-235B-A22B-Instruct, and Qwen3-30B-A3B-Thinking. As illustrated in Figure~\ref{fig:ablation_recall}, the PReSO workflow consistently achieves a significantly higher recall rate than TripTailor across all tested LLMs. Notably, when powered by GPT-4o, PReSO attains a recall of 42.26\%, a substantial increase over TripTailor’s 27.83\% (+14.43\%). These results validate that our hybrid, multi-dimensional retrieval mechanism is superior at capturing relevant environmental data, thereby providing a high-fidelity foundation for the subsequent generation of comprehensive itineraries.

\paragraph{Effectiveness of the CCoT Mechanism.}
Table~\ref{tab: ablation_agent_number} highlights the pivotal role of: (1) the competitive consensus mechanism, and (2) the impact of agent scaling on planning quality. Removing the CCoT arbitration protocol (w/o CCoT) leads to a marked decline in Macro Rationality (84.9\%) and the lowest Final Pass Rate (47.8\%), confirming that competitive consensus is essential for resolving the ``all-or-nothing'' conflicts inherent in multi-objective travel planning. Furthermore, our scaling analysis reveals a ``sweet spot'' in agent density: increasing from 3 to 4-6 agents yields gains in Rationality and Final Pass Rate, whereas further expansion to 10 agents results in diminishing returns, with Macro Rationality plateauing at 90.1\% and the Final Pass Rate slightly regressing. This suggests that while agent diversity is critical for capturing varied user preferences, a moderate-sized ensemble of specialized agents provides the optimal balance between comprehensive reasoning and stable arbitration.

\paragraph{Effectiveness of the Constraint-Gated RL.}
Table~\ref{tab: ablation_agent_number} demonstrates the necessity of: (1) RL, and (2) the constraint-gated reward. Compared to Constraint-Gated RL, direct refinement (w/o RL) shows a significant performance drop across all the metrics, which indicates the effectiveness of our RL process. Meanwhile, vanilla RL that directly adds hard and soft rewards also performs poorly, with a remarkable decline in Macro Rationality. This highlights the importance of our carefully designed gating mechanism in simultaneously optimizing common-sense constraints with personalized user preferences.

\section{Conclusion}
In this paper, we present TourPlanner, a comprehensive framework designed to enhance the quality of itineraries. Specifically, we first introduce the PReSO workflow, which leverages a three-branch recall mechanism and spatial clustering to prune candidate POIs, effectively constructing a spatially compact and information-rich candidate set. Furthermore, we propose CCoT, a multi-path reasoning paradigm that employs a multi-agent arbitration protocol to resolve conflicts and explore the feasible solution space, significantly improving the feasibility of generated itineraries. Finally, we implement a sigmoid-based gating mechanism within the reinforcement learning refinement stage; this curriculum-like approach ensures that soft-constraint rewards are dynamically prioritized only after hard constraints have been sufficiently satisfied. Experimental results demonstrate that TourPlanner achieves state-of-the-art performance, outperforming existing methods in both feasibility and alignment with user preferences.

\section*{Limitations}
TourPlanner has two main limitations that future research could address: (1) Due to the complexity and day-by-day iterative generation process of the CCoT mechanism, end-to-end optimization using RL is challenging. Although RL can be applied to single-day itinerary generation, defining an effective process reward function is difficult, as travel planning rewards typically depend on the overall quality of the entire trip. (2) The exploration of reward models in this study is limited, primarily adopting methodologies from previous works. Future research should focus on aligning the reward mechanism more closely with user preferences to achieve a higher surpassing rate.

\section*{LLM statement}
Large Language Models (LLMs) were employed for language refinement in this paper.
Specifically, we used them to polish grammar, improve clarity, and enhance the academic style of our writing.

\bibliography{custom}

\appendix
\newpage

\section{Details of GSPO}
\label{appendix: gspo}
GSPO \citep{zheng2025group} is a sequence-level policy optimization algorithm. GSPO introduces an importance ratio at the sequence level rather than at the token level. This design choice helps to exclude excessively off-policy samples from gradient estimation. The core optimization objective, given in Eq. \ref{eq: gspo}, involves two critical components: Group-based Advantage Estimation $\hat{A}i$ and the Sequence Importance Ratio $s_i(\theta)$. The latter quantifies the change in likelihood of an entire sequence under the new policy $\pi_{\theta}$ relative to the old policy $\pi_{\theta_{\text{old}}}$. To reduce variance, $s_i(\theta)$ incorporates length normalization via a factor of $\frac{1}{|y_i|}$:

\begin{equation}
\begin{aligned}
    s_i(\theta) &= \left( \frac{\pi_\theta(y_i|x)}{\pi_{\text{old}}(y_i|x)} \right)^{\frac{1}{|y_i|}} \\
    &= \exp \left( \frac{1}{|y_i|} \sum_{t=1}^{|y_i|} \log \frac{\pi_\theta(y_{i,t}|x, y_{i,<t})}{\pi_{\text{old}}(y_{i,t}|x, y_{i,<t})} \right).
\end{aligned}
\end{equation}

\section{Core Symbols and Definitions}
\label{app: symbol}
For clarity and consistency, Table~\ref{tab:notation} presents the principal notations and definitions adopted in this paper, covering the major variables and constructs.

\begin{table*}[ht!]
\centering
\captionsetup{justification=centering,font={small}}
\caption{Core symbols and definitions used throughout the paper.}
\label{tab:notation}
\renewcommand{\arraystretch}{1.15}
\resizebox{\textwidth}{!}{
\begin{tabular}{p{0.9cm} p{3.35cm} p{10.3cm}}
\toprule
\textbf{Symbol} & \textbf{Name} & \textbf{Definition}  \\
\midrule
\emph{Travel} & \\
$Q$ & User Query & A natural language input containing travel requirements and preferences. \\
$I$ & Itinerary & A structured travel plan including accommodations, transportation, restaurants, and POIs. \\
$\mathcal{A}$ & Agent Set & $\{A_1, A_2, \ldots, A_N\}$; set of reasoning agents instantiated for CCoT. \\
$I_i$ & Agent Identity & Describes the role and expertise of agent $A_i$ (e.g., culture, gourmet). \\
$O_i$ & Objective Function & Optimization goal pursued by agent $A_i$ (e.g., maximize cultural value). \\
$R_i$ & Priority Rules & Ranked preferences guiding the decision-making of agent $A_i$. \\
$P_{\text{base}, d}$ & Skeleton Itinerary & A base plan draft for Day $d$ focusing on logical geographic sequences. \\
$P_{i,d}$ & Daily Proposal & Day-$d$ itinerary refined by agent $A_i$ based on $P_{\text{base}, d}$ and its objective $O_i$. \\
$C_{1..d}$ & Consensus Plan & Aggregated plan from Day 1 to $d$, synthesized by \textit{Competition and Consensus Arbitration}. \\
$\bar{S}_i$ & Average Similarity & The mean semantic similarity between proposal $P_i$ and its peer proposals. \\
$w_i$ & Agent Weight & Diversity-based weight used to balance representativeness and specialized uniqueness. \\
$s_{i,j}$ & Peer Review Score & Numerical evaluation of $P_j$ by agent $A_i$ within $[-10, 10]$. \\
$T_{i,j}$ & Review Critique & Natural language feedback generated during the parallel peer review phase. \\\hline
\emph{RL} & \\
$R_{\text{hard}}$ & Hard Constraint Reward & Reward component based on Feasibility and Rationality metrics (e.g., within sandbox, information completeness). \\
$R_{\text{soft}}$ & Soft Constraint Reward & Reward component incorporating the Average Route Distance Ratio and a rating score generated by an advanced LLM.  \\
$\eta$ & Hard Constraint Score & Hyperparameter to adjust the degree of satisfaction of hard constraints. \\
$\alpha(\eta)$ & Gating Function & Sigmoid gating term regulating the contribution of soft rewards. \\
$\tau$ & Feasibility Threshold & Activation point in the gating function where hard constraints are met. \\
$k$ & Slope Coefficient & Controls the sharpness of the transition in $\alpha(\eta)$. \\
$\pi_\theta$ & Policy & Parameterized policy model of the RL agent. \\
$\hat{A}_i$ & Group Advantage & Normalized advantage estimation within GSPO. \\
$s_i(\theta)$ & Importance Ratio & Ratio between new and old policy likelihoods for trajectory $i$. \\
$\varepsilon$ & Clipping Parameter & Controls stability in policy optimization (Eq.~\ref{eq: gspo}). \\
$\mathcal{D}$ & Training Dataset & Set of pre-generated plans used for RL fine-tuning. \\
$G$ & Rollout Count & Number of trajectory samples per query during GSPO training. \\
\bottomrule
\end{tabular}
}
\end{table*}

\begin{table*}[ht!]
\centering
\captionsetup{justification=centering,font={small}}
\caption{TourPlanner hyperparameters. Duration denotes the required days for a travel itinerary.}
\label{tab:baseline_hyper}
\begin{tabular}{lc}
\toprule
Hyperparameters & Value \\
\hline
\emph{Multi-dimension POIs Recall} & \\
\quad Semantic similarity recall number & $3 \times \text{duration}$ \\
\quad POI recall total number & $9 \times \text{duration}$ \\
\hline
\emph{DBSCAN} & \\
\quad Minimum samples & 4 \\
\quad Epsilon $(\epsilon)$ & 1 \\
\quad Minimum cluster number & duration \\
\hline
\emph{CCoT} & \\
\quad Winner plans number ($k$) & 3 \\
\quad Diversity weight smoothing constant ($\epsilon$) & 0.01 \\
\hline
\emph{Constraint-Gated RL} & \\
\quad Constraint satisfaction ($\tau$) & $0.75$ \\
\quad Transition speed ($k$) & 28 \\
\bottomrule
\end{tabular}
\end{table*}

\section{Additional Experiment Details}
\subsection{Hyperparameter}
\label{app: hyper}
The hyperparameters of the method involved in this paper are summarized in Table~\ref{tab:baseline_hyper}.

\subsection{Details of Metrics}
\label{app: metrics}
In this section, we provide the details of the metrics used in our experiments.
\paragraph{Feasibility Pass Rate.}
This metric assesses the fundamental feasibility of a plan. A plan is deemed infeasible if the generated plan includes hallucinations, such as incorrect departure and return details or an inability to match POIs within the sandbox environment.

\paragraph{Rationality Pass Rate.}
This metric is assessed based on the five rationality criteria: Diverse Restaurants, Reasonable Meal Prices, Diverse Attractions, Appropriate Visit Duration, Appropriate Visit Time, and Defined Budget Limit. A plan must satisfy all five criteria to pass.

\paragraph{Average Route Distance Ratio.}
This metric evaluates the transportation efficiency of a plan. It is presented as a ratio of the average distance between consecutive POIs of the LLM-generated plan $D_{\text{avg}}^{\text{gen}}$ to that of the real plan $D_{\text{avg}}^{\text{real}}$. A lower ratio indicates a more efficient and compact route. The average distance for a single plan is calculated by first finding the average daily segment distance and then averaging this across all days of the trip:
\begin{equation}
    D_{\text{avg}} = \frac{\sum_{k=1}^{n_d} \left( \frac{\sum_{j=1}^{M_k-1} d_{j,j+1}^k}{M_k - 1} \right)}{n_d},
\end{equation}
where \( n_d \) is the total number of days in the itinerary, \( M_k \) is the number of POIs for day \( k \), and \( d_{j,j+1}^k \) is the distance between consecutive POIs on day \( k \).

\paragraph{Final Pass Rate.}
This metric integrates the above criteria: a plan must pass both Feasibility and Rationality assessments, and its total route length must not exceed 1.5 times the length of the reference plan.

\paragraph{Final Surpassing Rate.}
This metric assesses a model’s capability to match or exceed the personalization quality of human-created plans. It is evaluated using an ``LLM-as-a-Judge'' approach. In our paper, we use the Gemini-3-Pro model to compare the generated plan against the real plan.

\subsection{Details of Triptailor.}
\label{app: triptailor}
In contrast to methods that rely on dynamic online APIs for authentic POI data, which introduce variability and complicate fair evaluation, TripTailor provides a comprehensive, static dataset curated for complex itinerary planning. It integrates over 28,000 train schedules, 15,000 flight routes, 5,622 attractions, 89,000 hotels, and 422,000 restaurants, thereby offering a rich and structured basis for modeling multi-constraint travel scenarios. The dataset includes 3,145 training and 703 test samples, supporting robust evaluation of itinerary generation under diverse user requirements and spatial constraints.

\subsection{Baselines}
\label{app: baselines}
\begin{itemize}[leftmargin=10pt]
    \item \textbf{LLM Backbones:} We selected \href{https://openai.com/api/}{OpenAI GPT-4o} as the representative closed-source model. This is benchmarked against three recent and representative open-source models: Qwen3-235B-A22B-Instruct, Qwen3-30B-A3B-Thinking~\citep{yang2025qwen3}, and DeepSeek-R1~\citep{guo2025deepseek}. Notably, the latter two are designated thinking models.
    \item \textbf{Planning Approaches:} We compared our TourPlanner framework with three distinct planning approaches, namely, \textit{Direct Planning}, \textit{ReAct Planning}~\citep{zhang2024codeagent}, and the structured \textit{TripTailor Workflow}~\citep{wang2025triptailor}. The TripTailor workflow serves as a strong, non-LLM-centric baseline, which methodologically decomposes the planning process into sequential steps: identifying intercity transportation routes, using the LLM to rank and select attractions, generating an initial itinerary, integrating geographically proximate dining and accommodation options, and finally producing a comprehensive daily schedule.
\end{itemize}

\subsection{Reward Settings.}
\label{app: reward}

\subsubsection{Hard-Constraint Reward.}
The hard-constraint reward ($R_{\text{hard}}$) evaluates the fundamental viability of a travel plan. It is computed as the average of two components: Feasibility ($S_{\text{feas}}$) and Rationality ($S_{\text{rat}}$):
\begin{equation}
    R_{\text{hard}} = S_{\text{feas}} + S_{\text{rat}} = 2 \cdot \eta.
\end{equation}

\paragraph{Feasibility ($S_{\text{feas}}$).}
This component checks whether the plan contains hallucinations or incomplete information. It is the average of two binary indicators:
\begin{itemize}
    \item \textbf{Sandbox Validity ($I_{\text{sandbox}}\in\{0, 1\}$):} Checks if all entities (hotels, transportation, attractions, restaurants) exist within the TripTailor sandbox database.
    \item \textbf{Information Completeness ($I_{\text{comp}}\in\{0, 1\}$):} Verifies if essential details (prices, time schedules, modes of transport) are fully specified.
\end{itemize}
\begin{equation}
    S_{\text{feas}} = \frac{I_{\text{sandbox}} + I_{\text{comp}}}{2}.
\end{equation}

\paragraph{Rationality ($S_{\text{rat}}$).}
This component assesses the logical coherence of the itinerary based on four criteria:
\begin{itemize}
    \item \textbf{Restaurant Diversity ($I_{\text{rest}}\in\{0, 1\}$):} Ensures no restaurant is repeated across days.
    \item \textbf{Attraction Diversity ($I_{\text{attr}}\in\{0, 1\}$):} Ensures no attraction is visited more than once.
    \item \textbf{Duration Validity ($I_{\text{dur}}\in\{0, 1\}$):} Checks if the time spent at each attraction falls within the recommended duration range.
    \item \textbf{Visit Time Validity ($I_{\text{time}}\in\{0, 1\}$):} Verifies if the scheduled visit times align with the opening hours of the attractions.
\end{itemize}
\begin{equation}
    S_{\text{rat}} = \frac{I_{\text{rest}} + I_{\text{attr}} + I_{\text{dur}} + I_{\text{time}}}{4}.
\end{equation}

\subsubsection{Soft-Constraint Reward}
The soft-constraint reward ($R_{\text{soft}}$) incentivizes high-quality plans that optimize user preferences and logistical efficiency. It is a sum of three normalized scores:

\begin{equation}
    R_{\text{soft}} = S_{\text{budget}} + S_{\text{route}} + S_{\text{model}}.
\end{equation}

\begin{itemize}
    \item \textbf{Budget Score ($S_{\text{budget}}$):} Evaluates budget utilization and adherence. Let $C$ denote the total cost of the plan and $B$ the budget limit. The score encourages utilizing the available budget while linearly penalizing overspending:
    \begin{equation}
        S_{\text{budget}} = \begin{cases}
        \frac{C}{B}, & C \le B \\
        \max \left(0, 1 - \frac{C-B}{B}\right). & C > B
        \end{cases}
    \end{equation}
    \item \textbf{Route Efficiency Score ($S_{\text{route}}$):} Compares the average daily travel distance of the generated plan ($D_{\text{gen}}$) with that of an expert reference plan ($D_{\text{ref}}$). It is calculated as $S_{\text{route}} = \exp(-\max(0, \frac{D_{\text{gen}}}{D_{\text{ref}}} - 0.8))$, rewarding plans that maintain reasonable travel distances.
    \item \textbf{Preference Alignment Score ($S_{\text{model}}$):} We utilize a reward model to evaluate how well the plan satisfies implicit user preferences (e.g., ``relaxed pace'', ``cultural focus''). The reward model is trained following the dataset and methodology of TripTailor~\citep{wang2025triptailor}, with specific training configurations and the base model detailed in Appendix~\ref{app: Training Infrastructure}. Since positive and negative samples typically yield raw scores around $+10$ and $-10$ respectively, we apply a scaled hyperbolic tangent function to normalize these scores into a bounded range suitable for optimization:
\begin{equation}
    S_{\text{model}} = \tanh\left(\frac{\text{RM}(Q, I)}{6}\right),
\end{equation}
where $\text{RM}(Q, I)$ is the raw output of the Reward Model for query $Q$ and itinerary $I$. The scaling factor of 6 is empirically chosen to map the raw score distribution effectively onto the $[-1, 1]$ interval.
\end{itemize}

\subsubsection{Training Infrastructure}
\label{app: Training Infrastructure}
We employ Qwen2.5-3B-Instruct as the foundation model. Fine-tuning is conducted on 8 NVIDIA RTX 4090 GPUs with the following hyperparameters: a batch size of 4, a maximum sequence length of 4096 tokens, a learning rate of 1e-5, a weight decay of 0.01, 2 training epochs, and 2 gradient accumulation steps.

For RL fine-tuning, we utilize a cluster of 32 NVIDIA H800 GPUs. The setup includes a learning rate of 1e-6, a global batch size of 32 with a mini-batch size of 4, and 8 responses generated per prompt. We set the clipping range for GSPO to $[1 - \epsilon_\text{low}, 1 + \epsilon_\text{high}]$, where $\epsilon_\text{low}=0.0003, \epsilon_\text{high}=0.0004$, and limit the prompt and response lengths to 30,000 and 8,000 tokens, respectively. The training spans 3 epochs with dynamic batch sizing and sequence parallelism enabled.

\section{Prompts}
\label{app: prompts}
\subsection{PReSO workflow}
\subsubsection{Instruction for Demand Extraction}
\begin{tcolorbox}[
    breakable,
    colframe = gray,
    colback = gray!5!white,
    coltitle = white,
    coltext = black,
    fonttitle = \bfseries,
    boxrule = 1pt,
    arc = 2mm,
    width = \linewidth,
    left = 7pt,
    right = 7pt,
    top = 5pt,
    bottom = 5pt
]
\fontsize{8.5pt}{10pt}\selectfont
You are a travel assistant. When a user provides a travel query, extract and organize the information in the following structured format. If certain details (except `Other Requirements`) are not explicitly mentioned, infer them based on the user's query and general travel knowledge. Ensure each field is enclosed in square brackets `[]` for easy parsing. If no `Other Requirements` or `Restaurant Type` are mentioned, leave it blank. If the departure or return time is specified as ``morning'' assume it means ``early morning''\\
\\
\textbf{Structured Output Format:}\\
1. Departure Day: [Day of the Week]\\
2. Return Day: [Day of the Week]\\
3. Departure Time: [early morning/late morning/afternoon/evening]\\
4. Return Time: [early morning/late morning/afternoon/evening]\\
5. Duration: [Number of Days]\\
6. Departure City: [City Name]\\
7. Destination City: [City Name]\\
8. Other Requirements: [List of Requirements]\\
9. Budget: [Budget]\\
10. Restaurant Type: [Hot Pot/Fast Food/Northwestern Cuisine/Snacks/Buffet/Seafood/Pizza/Barbecue/Crayfish/Hainan Cuisine/Wontons and Dumplings/Sichuan Cuisine/Southeast Asian Cuisine/Jiangsu and Zhejiang Cuisine/Hunan Cuisine/Yunnan and Guizhou Cuisine/Porridge Shop/Other Delicacies/Rice Noodles/Korean Cuisine/Guangdong cuisine/Japanese Cuisine/Xinjiang Cuisine/Western Cuisine/Northeastern Cuisine/Malatang/Shandong Cuisine/Farmhouse Cuisine/Huaiyang Cuisine/Creative Cuisine/Vegetarian Cuisine/Jiangxi Cuisine/Chaoshan Cuisine/Anhui Cuisine/Taiwanese Cuisine/Tea Restaurant/Home-style Cooking/Hubei Cuisine/Beijing Cuisine/Fujian Cuisine/Guizhou Cuisine/Private Kitchen/Guangxi Cuisine/Hakka Cuisine/Tianjin Cuisine/Shanxi Cuisine/Henan Cuisine/Shaanxi Cuisine]\\
\\
\textbf{Example Query:}\\
``I am looking for a 4-day trip from Nanjing to Shenzhen, departing on Saturday early morning and returning on Tuesday afternoon, with a budget of ¥4000. I'm interested in exploring historical sites, cultural landmarks, scenic coastal parks, and relaxing natural retreats, along with enjoying diverse cuisines like seafood, Chaoshan, Hakka, and Guangdong dishes. The itinerary should be moderate in intensity, balancing guided exploration with some downtime.''\\
\\
\textbf{Expected Response:}\\
Departure Day: [Saturday]\\
Return Day: [Tuesday]\\
Departure Time: [early morning]\\
Return Time: [afternoon]\\
Duration: [4]\\
Departure City: [Nanjing]\\
Destination City: [Shenzhen]\\
Other Requirements: [exploring historical sites, cultural landmarks, scenic coastal parks and relaxing natural retreats]\\
Budget: [4000]\\
Reastaurant Type: [Seafood, Chaoshan Cuisine, Hakka Cuisine, Guangdong cuisine]\\
\\
\textbf{Now respond to the user query based on the examples provided above:}\\
\{user\_query\}
\end{tcolorbox}

\subsubsection{Instruction for Demand Inference}
\begin{tcolorbox}[
    breakable,
    colframe = gray,
    colback = gray!5!white,
    coltitle = white,
    coltext = black,
    fonttitle = \bfseries,
    boxrule = 1pt,
    arc = 2mm,
    width = \linewidth,
    left = 7pt,
    right = 7pt,
    top = 5pt,
    bottom = 5pt
]
\fontsize{8.5pt}{10pt}\selectfont
You are a travel assistant. Use the given statistics and the total budget to infer demands. Output exactly the two lines in the required format. No extra text.\\
\\
\textbf{Decision Rules}\\
- Hotel Cost (best within budget):\\
\hspace*{5mm} 1) Define nights N from the query, N = travel days - 1.\\
\hspace*{5mm} 2) Per-night hotel budget = (Budget $\times$ 0.55) / N.\\
\hspace*{5mm} 3) Categories priority: Luxury $>$ Upscale $>$ Midscale $>$ Economy.\\
\hspace*{5mm} 4) Choose the highest category c with Hotel Prices[c].min\_price $\leq$ per-night hotel budget.\\
\hspace*{5mm} 5) If none fits, choose the cheapest available category.\\
\\
- Meal Cost Range (allocated from total budget):\\
\hspace*{5mm} 1) Per-day meal budget = (Budget $\times$ 0.35) / N.\\
\hspace*{5mm} 2) Output a reasonable \textbf{integer range} based on budget and price statistics.\\
\hspace*{5mm} 3) Make the range slightly broader (not too narrow) to allow more restaurant options.\\
\hspace*{5mm} 4) The cost range can be adjust based on user's cuisin preference.\\
\\
Ensure each field is enclosed in square brackets `[]` for easy parsing. Organize the information in the following structured format:\\
\\
\textbf{Structured Output Format:}\\
1. Hotel Cost: [Luxury/ Upscale/ Midscale/ Economy]\\
2. Meal Cost Range: [Minimum Cost, Maximum Cost]\\
\\
\textbf{Example Response:}\\
Hotel Cost: [Midscale]\\
Meal Cost Range: [100,200]\\
\\
\textbf{Now respond to the user query based on the above information:}\\
\\
User Query:\\
\{user\_query\}\\
\\
Price Information:\\
Trasportation Prices: \{transportation\_info\}\\
Hotel Prices: \{hotel\_prices\_info\}\\
Restaurant Meal Prices: \{meal\_prices\_info\}\\
Budget: \{budget\}
\end{tcolorbox}

\subsection{CCoT}
\subsubsection{Instruction for Building Agents}
\begin{tcolorbox}[
    breakable,
    colframe = gray,
    colback = gray!5!white,
    coltitle = white,
    coltext = black,
    fonttitle = \bfseries,
    boxrule = 1pt,
    arc = 2mm,
    width = \linewidth,
    left = 7pt,
    right = 7pt,
    top = 5pt,
    bottom = 5pt
]
\fontsize{8.5pt}{10pt}\selectfont
You are a Chief Travel Planner. Analyze the user's travel query to identify core motives, constraints, and potential conflicts, then create a set of specialized \textbf{agent roles} in JSON ARRAY format. Each item must include fields:\\
1. ``agent\_id''\\
2. ``objective'' (measurable; e.g., ``minimize average leg distance (km)'', ``keep total cost $\leq$ budget'', ``maximize cultural-hours within opening hours'', ``ensure all meals within [min,max] CNY'')\\
3. ``priorities'' (ranked list of rules)\\
4. ``personality'' (short traits)\\
\\
\textbf{Example Response:}\\
{[}\\
\hspace*{4mm} \{ \\
\hspace*{8mm} ``agent\_id'': ``cultural\_scholar'',\\
\hspace*{8mm} ``objective'': ``Maximize the daily 'Cultural and Historical Experience' score (1-10)'',\\
\hspace*{8mm} ``priorities'': {[}``Visit museums $>$ 3 h'', ``World Heritage Sites'', ``Historic districts''{]},\\
\hspace*{8mm} ``personality'': ``Rigorous, inquisitive, dislikes commercialization''\\
\hspace*{4mm} \},\\
\hspace*{4mm} \{ \\
\hspace*{8mm} ``agent\_id'': ``foodie\_explorer'',\\
\hspace*{8mm} ``objective'': ``Maximize average meal quality score while keeping price $\leq$ 150 CNY'',\\
\hspace*{8mm} ``priorities'': {[}``Try regional cuisines'', ``Avoid chains'', ``Prefer authentic local spots''{]},\\
\hspace*{8mm} ``personality'': ``Curious, sociable''\\
\hspace*{4mm} \},\\
\hspace*{4mm} \{ \\
\hspace*{8mm} ``agent\_id'': ``budget\_manager'',\\
\hspace*{8mm} ``objective'': ``Keep total daily cost $\leq$ budget and maximize value'',\\
\hspace*{8mm} ``priorities'': {[}``Use public transport'', ``Choose economy restaurants''{]},\\
\hspace*{8mm} ``personality'': ``Pragmatic, cost-aware''\\
\hspace*{4mm} \}\\
{]}\\
\\
\textbf{Rules:}\\
1. Generate 4-6 agents depending on the complexity of the query. Fewer if simple, more if conflicting demands exist.\\
2. Each objective MUST include a measurable component (hours, CNY, km, count, etc.).\\
3. Avoid vague or filler goals such as ``relax'' or ``take it easy''. Instead, replace them with purposeful activities.\\
4. At least 80\% of daylight hours should have active or semi-active plans.\\
5. STRICT JSON only. No comments, no Markdown.\\
\\
\textbf{Now respond based on the user query:}\\
\{user\_query\}
\end{tcolorbox}

\subsubsection{Instruction for Building Per-Agent Day Plan}
\begin{tcolorbox}[
    breakable,
    colframe = gray,
    colback = gray!5!white,
    coltitle = white,
    coltext = black,
    fonttitle = \bfseries,
    boxrule = 1pt,
    arc = 2mm,
    width = \linewidth,
    left = 7pt,
    right = 7pt,
    top = 5pt,
    bottom = 5pt
]
\fontsize{8.5pt}{10pt}\selectfont
You are a Agent with the following profile: \{agent\_profile\}\\
Your task is to generate a detailed itinerary for [\{day\_label\}] that fulfills your own objectives and priorities while ensuring overall itinerary coherence, balance, and logical flow within the full trip plan. The output Format is a strict JSON object with:\\
1. agent\_id (eg. heritage\_historian)\\
2. day\_label (The specific day this plan is for, e.g., ``Day 1'')\\
3. daily\_cost (The total calculated cost for this day in CNY, including transportation (flights/trains for departure/return days), entrance fees, meals, and hotel expenses for all nights except the final return date)\\
4. plan (array of steps, where each step is an object with ``time'', ``activity type'', ``name'' and ``description'')\\
\\
\textbf{Example Response:}\\
\{ \\
\hspace*{4mm} ``agent\_id'': ``heritage\_historian'',\\
\hspace*{4mm} ``day\_label'': ``Day 1'',\\
\hspace*{4mm} ``daily\_cost'': 1250,\\
\hspace*{4mm} ``plan'': {[} \\
\hspace*{8mm} \{ \\
\hspace*{12mm} ``time'': ``07:45-09:15'',\\
\hspace*{12mm} ``activity type'': ``transportation'',\\
\hspace*{12mm} ``name'': ``CA8219'',\\
\hspace*{12mm} ``description'': ``Travel from Wuhan to Xi'an via flight CA8219.'' \\
\hspace*{8mm} \}, \dots \\
\hspace*{4mm} {]} \\
\}\\
\\
\textbf{Rules:}\\
1. \textbf{Source Integrity}: All attractions, restaurants, hotels, and transportations must be strictly selected only from their respective sections in the Given Information, not from any other descriptions or summaries.\\
2. \textbf{Traffic and Hotel Information}: \textbf{Day 1} begins with \textbf{arrival transportation}, then transfer to hotel and \textbf{check-in}. The \textbf{last day} ends with \textbf{hotel check-out} then transfer to airport/station and \textbf{return transportation}. The check-in and check-out must each be listed as independent activities (see example).\\
3. \textbf{Single Cluster Commitment}: Prefer attractions and restaurants within the same cluster to avoid long detours. Restaurants should be chosen near the preceding attraction, i.e., along the travel route.\\
4. \textbf{Visit Time and Duration}: Every visit must lie within \textbf{opening\_hours} and the visit duration must be between the lower and upper bound of \textbf{recommended duration}(0.5 day = 5 hours; 1 day = 10 hours), particularly must not be shorter than the minimum.\\
5. \textbf{Time and Sequence Constraint}:\\
\hspace*{5mm} All activities must be sequential and non-overlapping, with total active time between 7--8 hours (except for arrival/departure days). Activities \textbf{must not end later than 22:30}.\\
\hspace*{5mm} All consecutive activities must be connected by a \textbf{local\_transfer} activity, whose duration is estimated based on the distance between consecutive activities (at least 30 minutes), representing the travel between them. \textbf{No idle gap exceeding 1 hour}.\\
6. \textbf{Meal Enforcement}: Each day must contain lunch and dinner (except for arrival/departure days). Lunch must start between 11:00--14:00 and dinner between 17:00--20:00, and their start times must be at least 5 hours apart. Skipping lunch or dinner is not allowed; breakfast is not allowed.\\
7. \textbf{Diversity}: The same restaurant or attraction cannot be repeated within the same day or across different days, including those already used in previous days.\\
8. \textbf{Budget Guard}: Keep total cost within the user's expected range.\\
9. \textbf{Preference}: Balance cultural/natural/culinary exposure; prefer items that match the user's core motives.\\
10. It is not permissible for a half-day (morning or afternoon) to be completely empty, unless it is the departure day or the return day.\\
11. No activities will be scheduled 2 hours before flight departure and 1 hours before train departure.\\
12. \textbf{activity type} be one of the following: [``transportation'', ``check-in'', ``check-out'', ``sightseeing'', ``meal'', ``local\_transfer'']\\
\hspace*{5mm} - ``transportation'': Train or Flight travel between the departure and destination cities.\\
\hspace*{5mm} - ``check-in'' / ``check-out'': Applicable only to hotel accommodations.\\
\hspace*{5mm} - ``sightseeing'': Visiting landmarks, scenic spots, cultural sites, street/quarter, and attending performances, the locations should be selected from the attractions section of the \textbf{Given Information}.\\
\hspace*{5mm} - ``meal'': Meals at restaurants, cafes, or eateries (\textbf{excluding breakfast}), the locations should be selected from the restaurants section of the \textbf{Given Information}.\\
\hspace*{5mm} - ``local\_transfer'': Transfer between sequential activities within the same city (e.g., subway, bus, taxi, bicycle, or walking).\\
\\
\textbf{\#\# GIVEN\_INFORMATION}\\
\{city\_information\}\\
\\
\textbf{\#\# Previous Days Plan}\\
\{previous\_days\_plan\}\\
\\
\textbf{\#\# USER QUERY (Must align with it.)}\\
\{user\_query\}\\
\\
\textbf{\#\# Hard Rules:}\\
1. \textbf{Source Integrity}: All attractions, restaurants, hotels, and transportations must be strictly selected only from their respective sections in the Given Information, not from any other descriptions or summaries.\\
2. \textbf{Traffic and Hotel Information}: \textbf{Day 1} begins with \textbf{arrival transportation}, then transfer to hotel and \textbf{check-in}. The \textbf{last day} ends with \textbf{hotel check-out} then transfer to airport/station and \textbf{return transportation}. The check-in and check-out must each be listed as independent activities (see example).\\
3. \textbf{Single Cluster Commitment}: Prefer attractions and restaurants within the same cluster to avoid long detours. Restaurants should be chosen near the preceding attraction, i.e., along the travel route.\\
4. \textbf{Visit Time and Duration}: Every visit must lie within \textbf{opening\_hours} and the visit duration must be between the lower and upper bound of \textbf{recommended duration}(0.5 day = 5 hours; 1 day = 10 hours), particularly must not be shorter than the minimum.\\
5. \textbf{Time and Sequence Constraint}:\\
\hspace*{5mm} All activities must be sequential and non-overlapping, with total active time between 7--8 hours (except for arrival/departure days). Activities \textbf{must not end later than 22:30}.\\
\hspace*{5mm} All consecutive activities must be connected by a \textbf{local\_transfer} activity, whose duration is estimated based on the distance between consecutive activities (at least 30 minutes), representing the travel between them. \textbf{No idle gap exceeding 1 hour}.\\
6. \textbf{Meal Enforcement}: Each day must contain lunch and dinner (except for arrival/departure days). Lunch must start between 11:00--14:00 and dinner between 17:00--20:00, and their start times must be at least 5 hours apart. Skipping lunch or dinner is not allowed; breakfast is not allowed.\\
7. \textbf{Diversity}: The same restaurant or attraction cannot be repeated within the same day or across different days, including those already used in previous days.\\
8. \textbf{Budget Guard}: Keep total cost within the user's expected range.\\
9. It is not permissible for a half-day (morning or afternoon) to be completely empty, unless it is the departure day or the return day.\\
10. No activities will be scheduled 2 hours before flight departure and 1 hours before train departure.\\
11. \textbf{activity type} be one of the following: [``transportation'', ``check-in'', ``check-out'', ``sightseeing'', ``meal'', ``local\_transfer'']\\
\hspace*{5mm} - ``transportation'': Train or Flight travel between the departure and destination cities.\\
\hspace*{5mm} - ``check-in'' / ``check-out'': Applicable only to hotel accommodations.\\
\hspace*{5mm} - ``sightseeing'': Visiting landmarks, scenic spots, cultural sites, street/quarter, and attending performances, the locations should be selected from the attractions section of the \textbf{Given Information}.\\
\hspace*{5mm} - ``meal'': Meals at restaurants, cafes, or eateries (\textbf{excluding breakfast}), the locations should be selected from the restaurants section of the \textbf{Given Information}.\\
\hspace*{5mm} - ``local\_transfer'': Transfer between sequential activities within the same city (e.g., subway, bus, taxi, bicycle, or walking).\\
\\
Now response according to the GIVEN\_INFORMATION, USER QUERY, and Rules above:
\end{tcolorbox}

\subsubsection{Instruction for Peer Review}
\begin{tcolorbox}[
    breakable,
    colframe = gray,
    colback = gray!5!white,
    coltitle = white,
    coltext = black,
    fonttitle = \bfseries,
    boxrule = 1pt,
    arc = 2mm,
    width = \linewidth,
    left = 7pt,
    right = 7pt,
    top = 5pt,
    bottom = 5pt
]
\fontsize{8.5pt}{10pt}\selectfont
You are role-playing as a travel agent with this profile: \{reviewer\_agent\_profile\}\\
\\
\textbf{Task}: Review all competing plans. Judge each plan by your objectives/priorities.\\
\\
SCORE = BASELINE + PRIORITY\_FIT + BONUSES $-$ PENALTIES, then clamp to [-10, +10], integers only.\\
$\bullet$ \textbf{-10 only for a truly empty plan} (no real content).\\
$\bullet$ BASELINE = +2 (prevents universal negatives unless a plan is genuinely bad).\\
\\
\textbf{A) PRIORITY\_FIT (0--10):}\\
0=irrelevant, 3=weak, 6=good, 8=strong, 10=excellent alignment with your priorities.\\
\\
\textbf{B) BONUSES (0--6):}\\
+0--3 Spatial coherence (clustered routing, minimal detours)\\
+0--2 Diversity within theme (non-redundant POIs/meals)\\
+0--1 Budget/comfort fit (only if your profile cares)\\
\\
\textbf{C) PENALTIES} (use different discount based on the \textbf{severity}: Minor= -0.5, Major= -1, Critical= -1.5; sum all):\\
1. \textbf{Traffic and Hotel Information}: \textbf{Day 1} begins with \textbf{arrival transportation}, then transfer to hotel and \textbf{check-in}. The \textbf{last day} ends with \textbf{hotel check-out} then transfer to airport/station and \textbf{return transportation}.\\
2. \textbf{Time and Sequence Constraint}: All activities must be sequential and non-overlapping. \textbf{no idle gap $>$ 1h}; include \textbf{$\ge$15min transfer}.\\
3. \textbf{Meal Enforcement}: Each day must contain lunch and dinner (except for arrival/departure days). Lunch must start between 11:00--14:00 and dinner between 17:00--20:00, and their start times must be at least 5 hours apart.\\
\\
\textbf{Procedure} (mental; output only scores):\\
1) Empty check $\rightarrow$ if truly empty, score = -10.\\
2) Compute PRIORITY\_FIT, BONUSES, severity-weighted PENALTIES, then raw score = clamp\_round(B+F+Bon$-$Pen, [-10,+10]).\\
3) \textbf{Rank \& Stretch to force spread} (non-empty plans only):\\
\hspace*{4mm} a) Rank by raw score desc. Tie-breakers: fewer total penalties $\rightarrow$ higher PRIORITY\_FIT $\rightarrow$ shorter total idle gap $\rightarrow$ lexicographic agent\_id.\\
\hspace*{4mm} b) \textbf{Anchor}: lift best plan to at least \textbf{+7}, push worst (non-empty) to at most \textbf{-3} using linear rescale on the ranked list; keep interior order.\\
\hspace*{4mm} c) Ensure \textbf{$\ge$3 distinct integers} overall. If collisions remain, nudge neighbors by $\pm$1 (respecting [-10,+10]).\\
4) Guardrails:\\
$\bullet$ Never use -10 except for truly empty.\\
$\bullet$ Do not add commentary or unknown keys.\\
\\
Your Output Format must be a STRICT JSON object where keys are the ``agent\_id'' of the plans you reviewed.\\
For each plan, provide:\\
1. ``score'': integer [-10, 10]\\
2. ``critique'': A short string (max 30 words) explaining the main flaw or strength (e.g., ``Good route but ignores budget'', ``Perfect logical flow'').\\
\\
\textbf{Example Response:}\\
\{ \\
\hspace*{4mm} ``cultural\_scholar'': \{ \\
\hspace*{8mm} ``score'': 9, \\
\hspace*{8mm} ``critique'': ``Excellent cultural depth and logical route.'' \\
\hspace*{4mm} \},\\
\hspace*{4mm} ``budget\_manager'': \{ \\
\hspace*{8mm} ``score'': -2, \\
\hspace*{8mm} ``critique'': ``Violates opening hours and exceeds budget.'' \\
\hspace*{4mm} \} \\
\}\\
\\
Must align with user query: \{user\_query\}\\
\\
\textbf{--- ALL COMPETING PLANS ---}\\
\{plans\_joined\}
\end{tcolorbox}

\subsubsection{Instruction for Committee Arbitration}
\begin{tcolorbox}[
    breakable,
    colframe = gray,
    colback = gray!5!white,
    coltitle = white,
    coltext = black,
    fonttitle = \bfseries,
    boxrule = 1pt,
    arc = 2mm,
    width = \linewidth,
    left = 7pt,
    right = 7pt,
    top = 5pt,
    bottom = 5pt
]
\fontsize{8.5pt}{10pt}\selectfont
You are the COMMITTEE ARBITRATOR. Fuse multiple agents’ day plans into ONE realistic, feasible, and elegant itinerary for [\{day\_label\}].\\
Use ONLY items from Given Information (transport, attractions, restaurants, hotels). Do NOT invent names.\\
\textbf{daily\_cost}: The total calculated cost for this day in CNY, including transportation (flights/trains for departure/return days), entrance fees, meals, and hotel expenses for all nights except the final return date)\\
\\
\textbf{Example Response:}\\
\{ \\
\hspace*{4mm} ``day\_label'': ``Day 1'',\\
\hspace*{4mm} ``daily\_cost'': 1250,\\
\hspace*{4mm} ``plan'': {[} \\
\hspace*{8mm} \{ \\
\hspace*{12mm} ``time'': ``07:45-09:15'',\\
\hspace*{12mm} ``activity type'': ``transportation'',\\
\hspace*{12mm} ``name'': ``CA8219'',\\
\hspace*{12mm} ``description'': ``Travel from Wuhan to Xi'an via flight CA8219.'' \\
\hspace*{8mm} \}, \dots \\
\hspace*{4mm} {]} \\
\}\\
\\
\textbf{Rules:}\\
1. \textbf{Source Integrity}: All attractions, restaurants, hotels, and transportations must be strictly selected only from their respective sections in the \textbf{Given Information}, not from any other descriptions or summaries.\\
2. \textbf{Traffic and Hotel Information}: \textbf{Day 1} begins with \textbf{arrival transportation}, then transfer to hotel and \textbf{check-in}. The \textbf{last day} ends with \textbf{hotel check-out} then transfer to airport/station and \textbf{return transportation}. The check-in and check-out must each be listed as independent activities (see example).\\
3. \textbf{Single Cluster Commitment}: Prefer attractions and restaurants within the same cluster to avoid long detours. Restaurants should be chosen near the preceding attraction, i.e., along the travel route.\\
4. \textbf{Visit Time and Duration}: Every visit must lie within \textbf{opening\_hours} and the visit duration must be between the lower and upper bound of \textbf{recommended duration}(0.5 day = 5 hours; 1 day = 10 hours), particularly must not be shorter than the minimum.\\
5. \textbf{Time and Sequence Constraint}:\\
\hspace*{5mm} All activities must be sequential and non-overlapping, with total active time between 7--8 hours (except for arrival/departure days). Activities \textbf{must not end later than 22:30}.\\
\hspace*{5mm} There must be a \textbf{transfer time of at least 30 minutes}, estimated based on the distance between consecutive activities, and \textbf{no idle gap exceeding 1 hour}.\\
6. \textbf{Meal Enforcement}: Each day must contain lunch and dinner (except for arrival/departure days). Lunch must start between 11:00--14:00 and dinner between 17:00--20:00, and their start times must be at least 5 hours apart. Skipping lunch or dinner is not allowed; breakfast is not allowed.\\
7. \textbf{Diversity}: The same restaurant or attraction cannot be repeated within the same day or across different days, including those already used in previous days.\\
8. \textbf{Budget Guard}: Keep total cost within the user's expected range.\\
9. \textbf{Preference}: Balance cultural/natural/culinary exposure; prefer items that match the user's core motives.\\
10. It is not permissible for a half-day (morning or afternoon) to be completely empty, unless it is the departure day or the return day.\\
11. No activities will be scheduled 2 hours before flight departure and 1 hours before train departure.\\
12. \textbf{activity type} be one of the following: [``transportation'', ``check-in'', ``check-out'', ``sightseeing'', ``meal'', ``local\_transfer'']\\
\hspace*{5mm} - ``transportation'': Train or Flight travel between the departure and destination cities.\\
\hspace*{5mm} - ``check-in'' / ``check-out'': Applicable only to hotel accommodations.\\
\hspace*{5mm} - ``sightseeing'': Visiting landmarks, scenic spots, cultural sites, street/quarter, and attending performances, the locations should be selected from the attractions section of the \textbf{Given Information}.\\
\hspace*{5mm} - ``meal'': Meals at restaurants, cafes, or eateries (\textbf{excluding breakfast}), the locations should be selected from the restaurants section of the \textbf{Given Information}.\\
\hspace*{5mm} - ``local\_transfer'': Transfer between sequential activities within the same city (e.g., subway, bus, taxi, bicycle, or walking).\\
\\
\textbf{CRITICAL ROUTING INSTRUCTION:}\\
The candidate plans provided are variations of an optimized route. When creating the final plan:\\
1. \textbf{Maintain the geographic sequence} found in the best-rated candidate plan. Do not rearrange the order of locations arbitrarily, as this increases travel distance.\\
2. Select the specific POIs (Attractions/Restaurants) that maximize the consensus score, but keep them in the logical time slots.\\
3. If combining Plan A's morning and Plan B's afternoon, ensure the transition (Local Transfer) is geographically sensible.\\
\\
\textbf{\#\#\# PEER REVIEW INSIGHTS}\\
\{critique\_summary\}\\
\\
\textbf{\#\#\# INPUTS}\\
\\
Given Information: \{given\_info\_text\}\\
Proposals (JSON): \{plans\_joined\}\\
User Query: \{user\_query\}\\
Budget: \{budget\}\\
Is First Day?: \{is\_first\_day\}\\
Is Last Day?: \{is\_last\_day\}\\
Previous days' plan: \{previous\_days\_plan\}
\end{tcolorbox}

\subsection{Constraint-Gated RL}
\begin{tcolorbox}[
    breakable,
    colframe = gray,
    colback = gray!5!white,
    coltitle = white,
    coltext = black,
    fonttitle = \bfseries,
    title = Instruction for Plan Refinement,
    boxrule = 1pt,
    arc = 2mm,
    width = \linewidth,
    left = 7pt,
    right = 7pt,
    top = 5pt,
    bottom = 5pt
]
\fontsize{8.5pt}{10pt}\selectfont
You are a strict \textbf{travel plan validator \& fixer}.\\
\\
Your ONLY job: produce a fixed itinerary in the EXACT ``Output Format'' style shown below. Output the itinerary ONLY --- no preface, no analysis, no notes, no JSON, no bullet lists outside the format, no headings other than the time-slot lines, no code fences.\\
\\
\textbf{Inputs}:\\
- USER\_QUERY (budget/duration/interests)\\
- INITIAL\_PLAN (draft)\\
- GIVEN\_INFO (Clusters with Attractions / Restaurants / Hotels, plus Global Transportations)\\
\\
\textbf{Hard Constraints (all must pass)}\\
1. \textbf{Source Integrity}: All attractions, restaurants, hotels, and transportations must be strictly selected only from their respective sections in the Given Information, not from any other descriptions or summaries.\\
2. \textbf{Traffic and Hotel Information}: \textbf{Day 1} begins with \textbf{arrival transportation}, then transfer to hotel and \textbf{check-in}. The \textbf{last day} ends with \textbf{hotel check-out} then transfer to airport/station and \textbf{return transportation}.\\
3. \textbf{Single Cluster \& Route Continuity}: Prioritize attractions and restaurants within the same cluster to avoid long detours. The itinerary should follow a logical, sequential path, ensuring restaurants are chosen near the preceding attraction and situated directly along the travel route to maintain efficiency.\\
4. \textbf{Visit Time and Duration}: Every visit must lie within \textbf{opening\_hours} and the visit duration must be between the lower and upper bound of \textbf{recommended duration}(0.5 day = 5 hours; 1 day = 10 hours), particularly must not be shorter than the minimum.\\
5. \textbf{Time and Sequence Constraint}:\\
\hspace*{5mm} All activities must be sequential and non-overlapping. A minimum transfer time of 30 minutes is required between any two consecutive activities, estimated based on the distance between consecutive activities. Additionally, no idle interval may exceed one hour.\\
6. \textbf{Meal Enforcement}: Each day must include both lunch and dinner, except on arrival and departure days. If an attraction is described as offering food options (e.g., street food districts or food quarters), a visit scheduled during the lunch or dinner time window is considered to fulfill the corresponding meal requirement, and MUST NOT be represented as a separate meal activity.\\
7. \textbf{Diversity}: No restaurant or attraction may be visited more than once, either within the same day or across different days, including those already selected in previous days. Moreover, an attraction cannot be used separately for sightseeing and dining; any visit to the same physical location is counted as one single activity.\\
8. \textbf{Budget Guard}: Keep total cost within the user's expected range.\\
9. \textbf{Preference}: Balance cultural/natural/culinary exposure; prefer items that match the user's core motives.\\
10. No activities will be scheduled 2 hours before flight departure and 1 hours before train departure.\\
\\
\textbf{Repair Strategy (minimal)}\\
- Replace any non-compliant item with a compliant alternative (prefer same-cluster options).\\
- If replacement is impossible, drop the slot and tighten/shift neighboring items to avoid $>$90-minute gaps while staying within hours and recommended durations.\\
- Always enforce restaurant and attraction diversity, and time continuity.\\
\\
\textbf{Output Rules (STRICT)}\\
- Write ONLY the itinerary in the following style. No extra sections, no explanations, no JSON, no ``Output Format'' header, no code fences.\\
- Day title: bold line like \textbf{Day X Itinerary: <context>}.\\
- Each activity line uses the pattern: \texttt{\#\#\# HH:MM--HH:MM | <Title>}.\\
- After each activity, include the required details as short sentences and bullet points exactly as in the sample.\\
- Use \texttt{---} (three hyphens) as separators between activities.\\
- Use en dash `--' between times (e.g., 06:10--11:00).\\
- Keep prices and times exactly as in GIVEN\_INFO when referenced.\\
- Any activities at attraction must include a \textbf{Recommended Duration}.\\
\\
\textbf{Output Format (example to mimic EXACTLY)}\\
\textbf{Day 1 Itinerary: Wuhan to Xi'an}\\
\\
\#\#\# 07:45--09:15 | Travel to Xi'an\\
Start your journey with a \textbf{flight} on \textbf{CA8219} from \textbf{Wuhan} to \textbf{Xi'an Xianyang International Airport}. Depart at 07:45 and arrive at 09:15, ensuring a punctual and comfortable trip with a 99\% on-time performance.\\
- \textbf{Ticket Price}: ¥340\\
\\
---\\
\\
\#\#\# 10:00--10:30 | Check-in at Long March Mirasot Hotel\\
After arriving in Xi'an, check in at \textbf{Long March Mirasot Hotel}, an \textbf{Upscale hotel} with a 4.5/5 rating. Enjoy a comfortable stay at an average price of ¥424 per night. Guests have praised the hotel for its \textbf{excellent service} (9.0/10) and \textbf{pleasant environment} (8.0/10).\\
- \textbf{Average Price Per Night}: ¥424\\
\\
---\\
\\
\#\#\# 11:00--13:00 | Gao Family Mansion\\
Explore \textbf{Gao Family Mansion}, a century-old official compound located in Huimin Street. Discover the architectural art of the Ming and Qing Dynasties, traditional house couplets, brick carving art, and folk crafts like paper-cutting and shadow play.\\
- \textbf{Opening Hours}: 10:00--22:00\\
- \textbf{Entrance Fee}: ¥14.91\\
- \textbf{Recommended Duration}: 1-2 hours\\
\\
---\\
\\
\#\#\# 14:00--17:00 | Xi'an Expo Park\\
Visit \textbf{Xi'an Expo Park}, a garden park showcasing global horticultural art. Highlights include the Chang'an Tower, themed gardens, and rare plants from around the world. Perfect for leisurely walks and photography.\\
- \textbf{Opening Hours}: 09:00--22:00\\
- \textbf{Entrance Fee}: Free\\
- \textbf{Recommended Duration}: 2-4 hours\\
\\
---\\
\\
\#\#\# 18:00--19:30 | Dinner at Beijing Laoqianmen Roast Duck\\
Enjoy a delicious dinner at \textbf{Beijing Laoqianmen Roast Duck}, offering authentic Beijing cuisine. Savor the signature roast duck in a cozy setting with a 3.5/5 rating.\\
- \textbf{Average Price}: ¥105\\
- \textbf{Environment Rating}: 7.6/10\\
- \textbf{Service Rating}: 7.2/10
\end{tcolorbox}

\subsection{Evaluation}
\begin{tcolorbox}[
    breakable,
    colframe = gray,
    colback = gray!5!white,
    coltitle = white,
    coltext = black,
    fonttitle = \bfseries,
    title = Instruction for Final Surpassing Rate,
    boxrule = 1pt,
    arc = 2mm,
    width = \linewidth,
    left = 7pt,
    right = 7pt,
    top = 5pt,
    bottom = 5pt
]
\fontsize{8.5pt}{10pt}\selectfont
You are an AI assistant evaluating two travel plans based on following criteria:\\
\\
\textbf{\#\#\# Evaluation Criteria and Key Factors to Consider:}\\
$\bullet$ \textbf{Experiences}: Consider both variety and depth. While a diverse range of activities is beneficial, immersive and well-planned experiences that align closely with traveler interests should also be recognized.\\
$\bullet$ \textbf{Itinerary Intensity}: Evaluate how well the plan matches the traveler’s desired itinerary intensity (e.g., relaxed, moderate, packed). Balance activities with free time and ensure that no half-day (morning or afternoon) is completely empty, unless it is the arrival or departure day.\\
$\bullet$ \textbf{Cuisine}: Assess the suitability of dining choices to the traveler’s stated preferences, including cuisine category and alignment with budget.\\
$\bullet$ \textbf{Accommodations}: Evaluate the quality, comfort, and overall fit with the traveler’s stated preferences, including accommodation category and budget range.\\
$\bullet$ \textbf{Transportation}: Assess the practicality of transportation options with a focus on departure and return times, convenience, cost, and suitability for the traveler’s preferences.\\
$\bullet$ \textbf{Total Budget Consideration}: Staying within the budget is essential, but an itinerary that justifies slightly higher costs through premium experiences is viewed positively, whereas strict cost-cutting at the expense of premium experiences is seen as unfavorable.\\
$\bullet$ \textbf{Traffic and Hotel Information}: \textbf{Day 1} begins with \textbf{arrival transportation}, then transfer to hotel and \textbf{check-in}. The \textbf{last day} ends with \textbf{hotel check-out} then transfer to airport/station and \textbf{return transportation}.\\
$\bullet$ \textbf{Time and Sequence Constraint}: All activities must be sequential and non-overlapping. Include a transfer time based on the distance between consecutive activities, and ensure no idle gap exceeds 2 hours.\\
$\bullet$ \textbf{Meal Enforcement}: Each full day must include lunch and dinner at appropriate times. Their start times must be at least 5 hours apart. Skipping lunch or dinner is not allowed, and breakfast should not be included.\\
$\bullet$ \textbf{Diversity}: The same restaurant or attraction cannot be repeated within the same day or across different days, including those already used in previous days.\\
\\
\textbf{\#\#\# Scoring Scale (Out of 5)}\\
\textbf{5 (Excellent)}: The itinerary \textbf{exceeds expectations}, perfectly aligning with all user preferences. It offers unique, tailored experiences and exceptional value, ensuring a memorable and personalized journey.\\
\textbf{4 (Good)}: The itinerary \textbf{largely meets} the user’s needs, showing a strong level of personalization and value. However, there may be minor gaps in specific preferences or opportunities for deeper engagement that could enhance the overall experience.\\
\textbf{3 (Average)}: The itinerary \textbf{partially satisfies} the user’s query, incorporating some preferences but missing key elements in important areas. It fulfills basic requirements but lacks depth, creativity, or engagement in activities, cultural insights, or personalization, resulting in a feeling of generality and mediocrity.\\
\textbf{2 (Poor)}: The itinerary \textbf{barely meets} expectations, with significant gaps in personalization and relevance. Most elements do not align well with the user’s stated preferences, leading to a less enjoyable and uninspired experience.\\
\textbf{1 (Very Poor)}: The itinerary \textbf{fails to address} the user's query entirely, displaying no relevance to stated preferences. It is completely generic, offering little to no value or consideration for the user's unique needs and interests.\\
\\
\textbf{\#\#\# Output format:}\\
\textbf{\#\#\#\# Comparative Analysis:}\\
{[}Please analyze each plan first and then provide a rating in JSON format. Based on the Evaluation Criteria and Key Factors to Consider, provide a detailed comparative analysis of how well each plan meets the traveler’s preferences and the overall quality of each plan, explaining their strengths and weaknesses. Focus your evaluation on the actual schedule (the 'what' and 'when' of each activity). {]}\\
\\
\textbf{\#\#\#\# Scoring Results:}\\
\{ \\
\hspace*{4mm} ``Personalization Evaluation'': \{ \\
\hspace*{8mm} ``Scores'': \{ \\
\hspace*{12mm} ``Plan A'': X,\\
\hspace*{12mm} ``Plan B'': Y \\
\hspace*{8mm} \} \\
\hspace*{4mm} \} \\
\}\\
\\
\textbf{\#\#\# Input}\\
$\bullet$ \textbf{Query}: \{query\}\\
$\bullet$ \textbf{Plan A}: \{plan\_a\}\\
$\bullet$ \textbf{Plan B}: \{plan\_b\}
\end{tcolorbox}

\section{Case Study}
\label{app: case}

Table~\ref{tab:case_ccot_rl} presents an example of a user query and the corresponding planning results generated by TourPlanner.

\begin{table*}[!t]
    \centering
    \caption{An example of user query and travel planning results.}
    \fontsize{9pt}{11pt}\selectfont
    \begin{tabular}{p{0.98\linewidth}}
    \midrule
        \rowcolor{gray!20}\textbf{User query} \\
    \midrule
        I am looking for a 4-day trip from Wuhan to Xi'an, departing on Thursday morning and returning on Sunday evening, with a budget of ¥4100. I’m interested in exploring historical landmarks, cultural museums, scenic parks, and architectural marvels, along with enjoying diverse cuisines like barbecue, hot pot, Korean, and Japanese dishes. The itinerary should be moderate in intensity, balancing guided exploration with some downtime. \\
    \midrule
        \rowcolor{gray!20}
        \textbf{Model Output by TourPlanner w/o RL}\\
    \midrule
        \blue{**Day 1 Itinerary: Wuhan to Xi'an **} \#\#\# 07:45-09:20 | Travel to Xi'an. Start your journey with a **flight** on **CA8219** from **Wuhan** to **Xi'an** - **Ticket Price**: ¥340 \#\#\# 10:15-10:45 | Check-in at Kunyi Hotel, a **Upscale hotel** with a 5.0 rating. Enjoy a comfortable stay at an average price of 387.0 per night.  - **Average Price Per Night**: 387.0 \#\#\# 11:05-12:05  | Beijing Zhengyangmen Roast Duck Restaurant. Lunch at Beijing Zhengyangmen Roast Duck Restaurant (¥121.5), a Beijing cuisine restaurant renowned for its signature roast duck. The meal starts at 11:05, well within the lunch window (11:00–14:00), and the restaurant is located in Cluster 1 near cultural attractions. - **Cuisine Type**: Beijing Cuisine  - **Average Cost**: 121.5 \#\#\# 12:30-17:30 | Visit Shaanxi History Museum **Feature**: Explore the exhibition hall; a hidden gem for kids.; MuseumsTraveling through time among the vast and precious cultural relics. - **Entrance Fee**: 0.0 - **Recommended Duration**: 0.5-1 day - **Opening Hours**: Open at  09:00-17:30(Admission ends at 16:00) \#\#\# 17:50-19:20  | Haocheng Zhen Yangcheng Lake Hairy Crab. Dinner at Haocheng Zhen Yangcheng Lake Hairy Crab (¥120), a crayfish specialty restaurant offering a popular local street food experience. The meal starts at 17:50, within the dinner window (17:00–20:00), and is located near Jianfu Temple and Xi'an Museum in Cluster 1, ensuring minimal transfer time after the museum visit. - **Cuisine Type**: Crayfish  - **Average Cost**: 120.0  **Total Daily Cost**: ¥968.5 \\
        \blue{**Day 2 Itinerary **} \#\#\# 08:30-11:30 | Visit Huaqing Palace **Feature**: Get Close to Nature; Explore the Exhibition Hall; Springs; Historical SitesThe place where Emperor Xuanzong of Tang and Consort Yang bathed together. - **Entrance Fee**: 120.0 - **Recommended Duration**: 2-3 hours - **Opening Hours**: Opens at 07:00-18:00(Tickets available until 17:30; admission ends at 17:40) \#\#\# 12:15-13:45  | Pier Story Hot Pot. Lunch at Pier Story Hot Pot (¥120), a hot pot restaurant known for its flavorful broths and authentic Xi'an-style dining experience. The meal starts at 12:15, within the designated lunch window (11:00–14:00), and is located near cultural sites such as the Xi'an Incident Site and Haitang Soup. The 1.5-hour duration allows for a relaxed dining experience following the morning tour. - **Cuisine Type**: Hot Pot  - **Average Cost**: 120.0 \#\#\# 14:30-16:30 | Visit Lishan **Feature**: Mountains - **Entrance Fee**: 20.0 - **Recommended Duration**: 2-3 hours - **Opening Hours**: Opens at 07:00-19:00(Admission ends at 16:30) \#\#\# 17:15-18:45  | Weng Kee Seafood. Dinner at Weng Kee Seafood (¥124.5), a seafood restaurant offering fresh coastal flavors with a modern twist. The meal starts at 17:15, within the dinner window (17:00–20:00), and is located near The Everlasting Regret Show and Jiulong Lake. With a rating of 4.0 and strong product quality, it provides a refined culinary contrast to local Xi'an fare. - **Cuisine Type**: Seafood  - **Average Cost**: 124.5  **Total Daily Cost**: ¥771.5 \\
        \blue{**Day 3 Itinerary **} \#\#\# 09:00-13:00 | Visit The Terracotta Army and Qin Shi Huang Mausoleum Museum **Feature**: Historical sites; Museum; Historical SitesThe Terracotta Army of Emperor Qin Shi Huang - **Entrance Fee**: 120.0 - **Recommended Duration**: 4-5 hours - **Opening Hours**: Open at 08:30-18:00(Admission ends at 16:30) \#\#\# 13:50-14:50  | Dalongyi hot pot. Lunch at Dalongyi hot pot (¥123), a hot pot restaurant in Cluster 0 with a 3.5 rating, known for its authentic Shaanxi-style broth and strong service and environment scores. The meal starts at 13:50, within the lunch window (11:00–14:00), and is geographically convenient after the museum visit. - **Cuisine Type**: Hot Pot  - **Average Cost**: 123.0 \#\#\# 15:40-17:10 | Visit Xi'an City Wall **Feature**: Kid-friendly Hidden Gems; Historical Ruins; Military Sites; Historical Sites; Nighttime sightseeingThe historical witness of the Thirteen Dynasties' ancient capitals. - **Entrance Fee**: 54.0 - **Recommended Duration**: 1-4 hours - **Opening Hours**: Opens at 08:00-22:00(Tickets available until 22:00; admission ends at 22:00) \#\#\# 17:50-18:50  | Tangyuan · Xiaonanyang Jiangxi Restaurant. Dinner at Tangyuan · Xiaonanyang Jiangxi Restaurant (¥127.5), a Jiangxi cuisine restaurant in Cluster 3 with a moderate rating of 3.0. The meal starts at 17:50, within the dinner window (17:00–20:00), and is more than 5 hours after lunch. It provides culinary diversity by offering regional flavors distinct from the repeated hot pot meals of previous days. - **Cuisine Type**: Jiangxi Cuisine  - **Average Cost**: 127.5  **Total Daily Cost**: ¥811.5 \\
        \blue{**Itinerary for the Last Day of the Trip**} \#\#\# 12:15-13:15  | Alley old hot pot. Lunch at Alley old hot pot (¥130.5), a local hot pot restaurant in Cluster 3 with a 3.5 rating. The meal starts at 12:15, within the lunch window (11:00–14:00), and fulfills the culinary priority of experiencing authentic hot pot, a must-try Xi'an dining style. This family-run eatery offers a street food–style atmosphere near natural attractions, aligning with the user's interest in diverse regional cuisines. - **Cuisine Type**: Hot Pot  - **Average Cost**: 130.5 \#\#\# 13:45-14:45 | Visit Emperor Qinshihuang's Mausoleum Site Museum **Feature**: Historical Sites; Cemeteries - **Entrance Fee**: 0.0 - **Recommended Duration**: 1-2 hours - **Opening Hours**: Opens at 08:30-18:00(Admission ends at 16:30) \#\#\# 15:05-15:35 | Hotel Check-Out. Check out from your hotel and store your luggage if needed. Ensure you have all your belongings before heading to the train station/airport. \#\#\# 17:35-19:20 | Return to Xi'an with a **flight** on **CZ3890** from **Xi'an** to **Wuhan** - **Ticket Price**: ¥500 **Total Daily Cost**: ¥630.5" \\

    \bottomrule
    \end{tabular}
    \label{tab:case_ccot_rl}
\end{table*}

\begin{table*}[!t]
    \centering
    \ContinuedFloat
    \caption{An example of user query and travel planning results (Continued).}
    \fontsize{9pt}{11pt}\selectfont
    \begin{tabular}{p{0.98\linewidth}}
    \midrule
        \rowcolor{gray!20}
        \textbf{Model Output by TourPlanner}\\
    \midrule
        \blue{**Day 1 Itinerary: Wuhan to Xi'an **} \#\#\# 07:45–09:20 | Travel to Xi'an. Begin your 4-day trip by boarding flight **CA8219** from **Wuhan** to **Xi'an Xianyang International Airport**. This morning flight departs at 07:45 and arrives at 09:20, providing a smooth start to your journey with an excellent on-time record. **Ticket Price**: ¥340 --- \#\#\# 10:00–10:30 | Check-in at Kunyi Hotel. Upon arrival in Xi'an, check in at **Kunyi Hotel**, an **Upscale hotel** with a perfect 5.0/5 rating. Located conveniently in Cluster 1, it is highly praised for its exceptional service (9.1/10) and pristine environment (9.0/10). **Average Price Per Night**: ¥387 --- \#\#\# 11:00–12:00 | Lunch at Beijing Zhengyangmen Roast Duck Restaurant. Enjoy your first meal at **Beijing Zhengyangmen Roast Duck Restaurant**, specializing in authentic Beijing cuisine. Savor their renowned signature roast duck in a comfortable setting located near major cultural landmarks. **Average Price**: ¥121.5 **Environment Rating**: 6.9/10 **Service Rating**: 6.3/10 --- \#\#\# 12:30–17:30 | Shaanxi History Museum. Explore the **Shaanxi History Museum**, China's first large-scale modern national museum. Known as the "Treasury of China," it houses over 370,000 relics ranging from prehistoric tools to Tang dynasty murals, showcasing the essence of ancient Chinese civilization. **Opening Hours**: 09:00–17:30 **Entrance Fee**: Free **Recommended Duration**: 0.5-1 day --- \#\#\# 18:00–19:30 | Dinner at Haocheng Zhen Yangcheng Lake Hairy Crab. For dinner, visit **Haocheng Zhen Yangcheng Lake Hairy Crab**, a popular spot for crayfish and local seafood. Located in the heart of the city, it offers a vibrant dining atmosphere perfect for trying regional specialties. **Average Price**: ¥120 **Environment Rating**: 6.9/10 **Service Rating**: 6.9/10 \\
        \blue{**Day 2 Itinerary**} \#\#\# 09:00–12:00 | Huaqing Palace. Head to Lintong to visit **Huaqing Palace**, a famous royal garden and the site of the legendary love story between Emperor Xuanzong and Consort Yang. Explore the ancient bathing pools and beautifully preserved Tang dynasty architecture. **Opening Hours**: 07:00–18:00 **Entrance Fee**: ¥120 **Recommended Duration**: 2-3 hours --- \#\#\# 12:30–14:00 | Lunch at Pier Story Hot Pot. Enjoy a flavorful lunch at **Pier Story Hot Pot**, known for its authentic spicy broths and high-quality ingredients. This restaurant offers a traditional Xi'an hot pot experience right near the historical sites of Lintong. **Average Price**: ¥120 **Environment Rating**: 7.8/10 **Service Rating**: 7.5/10 --- \#\#\# 14:30–17:30 | Lishan. Spend your afternoon at **Lishan**, a scenic mountain adjacent to Huaqing Palace. Highlights include the historic Beacon Tower and the Bingjian Pavilion, offering lush forest paths and panoramic sunset views over Lintong. **Opening Hours**: 07:00–19:00 **Entrance Fee**: ¥20 **Recommended Duration**: 2-3 hours --- \#\#\# 18:00–19:30 | Dinner at Dalongyi hot pot. Experience another dimension of local flavors at **Dalongyi hot pot**. This well-regarded establishment provides a lively environment and attentive service, ensuring a satisfying end to your day in the Lintong district. **Average Price**: ¥123 **Environment Rating**: 7.8/10 **Service Rating**: 7.6/10 \\
        \blue{**Day 3 Itinerary**} \#\#\# 08:30–13:30 | The Terracotta Army and Qin Shi Huang Mausoleum Museum. Discover the world-renowned **Terracotta Army**, an archaeological marvel featuring thousands of life-sized pottery figurines. Visit Pits 1, 2, and 3 to witness the incredible military formation and the exquisite bronze chariots of the Qin Dynasty. **Opening Hours**: 08:30–18:00 **Entrance Fee**: ¥120 **Recommended Duration**: 4-5 hours --- \#\#\# 14:00–15:30 | Lunch at Weng Kee Seafood. Dine at **Weng Kee Seafood**, which offers a variety of fresh seafood dishes with high product ratings. Its location in the Lintong area makes it a convenient stop after your morning at the museum. **Average Price**: ¥124.5 **Environment Rating**: 6.6/10 **Service Rating**: 6.8/10 --- \#\#\# 16:30–19:30 | Xi'an City Wall. Return to the city center to walk or cycle atop the **Xi'an City Wall**, the most well-preserved ancient city wall in China. Enjoy the architectural grandeur of the Ming Dynasty and unique views of both the old and new parts of the city. **Opening Hours**: 08:00–22:00 **Entrance Fee**: ¥54 **Recommended Duration**: 1-4 hours --- \#\#\# 20:00–21:30 | Dinner at Tangyuan · Xiaonanyang Jiangxi Restaurant. For dinner, try something different at **Tangyuan · Xiaonanyang Jiangxi Restaurant**. This establishment offers regional Jiangxi cuisine, providing a diverse culinary contrast to the local Shaanxi dishes enjoyed earlier. **Average Price**: ¥127.5 **Environment Rating**: 7.1/10 **Service Rating**: 6.9/10 \\
        \blue{**Itinerary for the Last Day of the Trip**} \#\#\# 09:00–12:00 | Xi'an Museum. Visit the **Xi'an Museum**, which integrates historical relics, the Small Wild Goose Pagoda, and beautiful urban gardens. Learn about the evolution of Xi'an through its themed exhibition "Ancient Capital Xi'an" and the impressive model of the city walls. **Opening Hours**: 09:00–17:30 **Entrance Fee**: Free **Recommended Duration**: 2-3 hours  --- \#\#\# 12:30–14:00 | Lunch at Alley old hot pot. Savor your final meal in Xi'an at **Alley old hot pot**. This local favorite provides a street-food style atmosphere and authentic broths, fulfilling your preference for diverse hot pot experiences before heading home. **Average Price**: ¥130.5 **Environment Rating**: 7.4/10 **Service Rating**: 7.3/10 --- \#\#\# 14:30–15:00 | Check-out at Kunyi Hotel. Return to the hotel to complete your **check-out** at **Kunyi Hotel**. Ensure all your belongings are packed before your transfer to the airport for your return flight to Wuhan. **Average Price Per Night**: ¥387 --- \#\#\# 17:35–19:20 | Travel to Wuhan. Conclude your trip with return flight **CZ3890** from **Xi'an** to **Wuhan**. The flight departs at 17:35 and arrives at 19:20, allowing you to return home safely after an enriching 4-day exploration of Xi'an's heritage. **Ticket Price**: ¥500 \\

    \bottomrule
    \end{tabular}
    \label{tab:case_ccot_rl}
\end{table*}

\end{document}